\documentclass{article}

\usepackage{microtype}
\usepackage{graphicx}
\usepackage{subcaption}
\usepackage{booktabs}

\usepackage{hyperref}

\usepackage[accepted]{icml2026}

\usepackage{amsmath}
\usepackage{amssymb}
\usepackage{mathtools}
\usepackage{amsthm}
\usepackage{multicol,multirow}
\usepackage{colortbl}
\usepackage{xcolor}
\usepackage[most]{tcolorbox}

\usepackage[capitalize,noabbrev]{cleveref}

\theoremstyle{plain}

\theoremstyle{definition}

\theoremstyle{remark}

\definecolor{SkillBlue}{RGB}{0, 85, 150}
\definecolor{SkillBlueLight}{RGB}{220, 235, 252}
\definecolor{SkillGreen}{RGB}{0, 150, 130}
\definecolor{SkillGreenLight}{RGB}{220, 245, 240}
\definecolor{SkillOrange}{RGB}{230, 120, 40}
\definecolor{SkillOrangeLight}{RGB}{255, 243, 224}
\definecolor{SkillGray}{RGB}{100, 100, 100}
\definecolor{SkillGrayLight}{RGB}{245, 245, 245}
\definecolor{BestShade}{RGB}{255, 239, 188}
\definecolor{SecondShade}{RGB}{221, 238, 255}
\newcommand{\best}[1]{\cellcolor{BestShade}\textbf{#1}}
\newcommand{\second}[1]{\cellcolor{SecondShade}#1}

\newtcolorbox{skillbox}[1][Skill]{
    enhanced,
    breakable,
    colback=SkillGreenLight,
    colframe=SkillGreen,
    fonttitle=\bfseries\color{white},
    title=#1,
    coltitle=white,
    colbacktitle=SkillGreen,
    boxrule=1pt,
    arc=3pt,
    left=8pt,
    right=8pt,
    top=6pt,
    bottom=6pt,
    before skip=10pt,
    after skip=10pt,
}

\newtcolorbox{deleteskillbox}[1][Skill]{
    enhanced,
    breakable,
    colback=SkillOrangeLight,
    colframe=SkillOrange,
    fonttitle=\bfseries\color{white},
    title=#1,
    coltitle=white,
    colbacktitle=SkillOrange,
    boxrule=1pt,
    arc=3pt,
    left=8pt,
    right=8pt,
    top=6pt,
    bottom=6pt,
    before skip=10pt,
    after skip=10pt,
}

\newtcolorbox{skipskillbox}[1][Skill]{
    enhanced,
    breakable,
    colback=SkillGrayLight,
    colframe=SkillGray,
    fonttitle=\bfseries\color{white},
    title=#1,
    coltitle=white,
    colbacktitle=SkillGray,
    boxrule=1pt,
    arc=3pt,
    left=8pt,
    right=8pt,
    top=6pt,
    bottom=6pt,
    before skip=10pt,
    after skip=10pt,
}

\newtcolorbox{promptbox}[1][Prompt]{
    enhanced,
    breakable,
    colback=SkillBlueLight,
    colframe=SkillBlue,
    fonttitle=\bfseries\color{white},
    title=#1,
    coltitle=white,
    colbacktitle=SkillBlue,
    boxrule=1pt,
    arc=3pt,
    left=8pt,
    right=8pt,
    top=6pt,
    bottom=6pt,
    before skip=10pt,
    after skip=10pt,
}

\definecolor{MotivBorder}{RGB}{70,70,70}
\definecolor{MotivFill}{RGB}{245,248,252}

\icmltitlerunning{Learning Design Skills as Memory Policies for Agentic Photonic Inverse Design}

\begin{document}

\twocolumn[
\icmltitle{Learning Design Skills as Memory Policies for 
           Agentic Photonic Inverse Design}

\icmlsetsymbol{equal}{*}

\begin{icmlauthorlist}
\icmlauthor{Shengchao Chen}{uts}
\icmlauthor{Ting Shu}{szu}
\icmlauthor{Sufen Ren}{hnu}
\end{icmlauthorlist}

\icmlaffiliation{uts}{AAII, University of Technology Sydney}
\icmlaffiliation{szu}{School of Artificial Intelligence, Shenzhen University}
\icmlaffiliation{hnu}{School of Information and Communication Engineering, Hainan University}

\icmlcorrespondingauthor{Shengchao Chen}{shengchao.chen.uts@gmail.com}

\icmlkeywords{Photonic Crystal Fiber, Inverse Design, Memory-Augmented Agents, Reinforcement Learning, AI for Physics}

\vskip 0.3in
]

\printAffiliationsAndNotice{}  % no special notice (required even if empty)

\begin{abstract}
Photonic crystal fiber (PCF) inverse design remains challenging because candidate geometries must satisfy coupled optical targets under expensive electromagnetic simulation. Existing pipelines improve surrogate prediction or one-shot parameter recommendation, but they do not accumulate reusable design knowledge across iterative trials. We formulate PCF inverse design as a memory-policy learning problem and propose SkillPCF, a closed-loop agent framework that combines a physics-guided memory skill bank, reinforcement-learned skill selection, and simulator-grounded skill evolution. We further construct a real-world dataset with 479 expert interaction traces (2,507 spans) and 553 memory-dependent evaluation queries covering dispersion engineering, loss optimization, and multi-objective design. Experiments across multiple LLM backbones and classical baselines show that SkillPCF achieves stronger design-quality and efficiency trade-offs under practical simulation budgets, demonstrating the effectiveness of our proposed memory-skill learning paradigm for physics-aware PCF inverse design.
\end{abstract}

\section{Introduction}
\label{sec:intro}

Photonic crystal fibers (PCFs) offer a wide design space for controlling dispersion, confinement loss, modal properties, and nonlinear response through microstructured cladding geometry \cite{Russell_2003,Knight_1996,Birks_1997,markos2017hybrid}. This flexibility is also the source of the core inverse-design difficulty: practical designs often require balancing multiple coupled targets over a high-dimensional parameter space, where each candidate evaluation may invoke expensive finite-element or finite-difference time-domain (FDTD) simulation~\cite{chen2025pcf}. As a result, optimization loops of structure design can become simulation-bound and brittle under changing constraints.

\begin{figure*}[tbh]
    \centering
    \includegraphics[width=.95\textwidth]{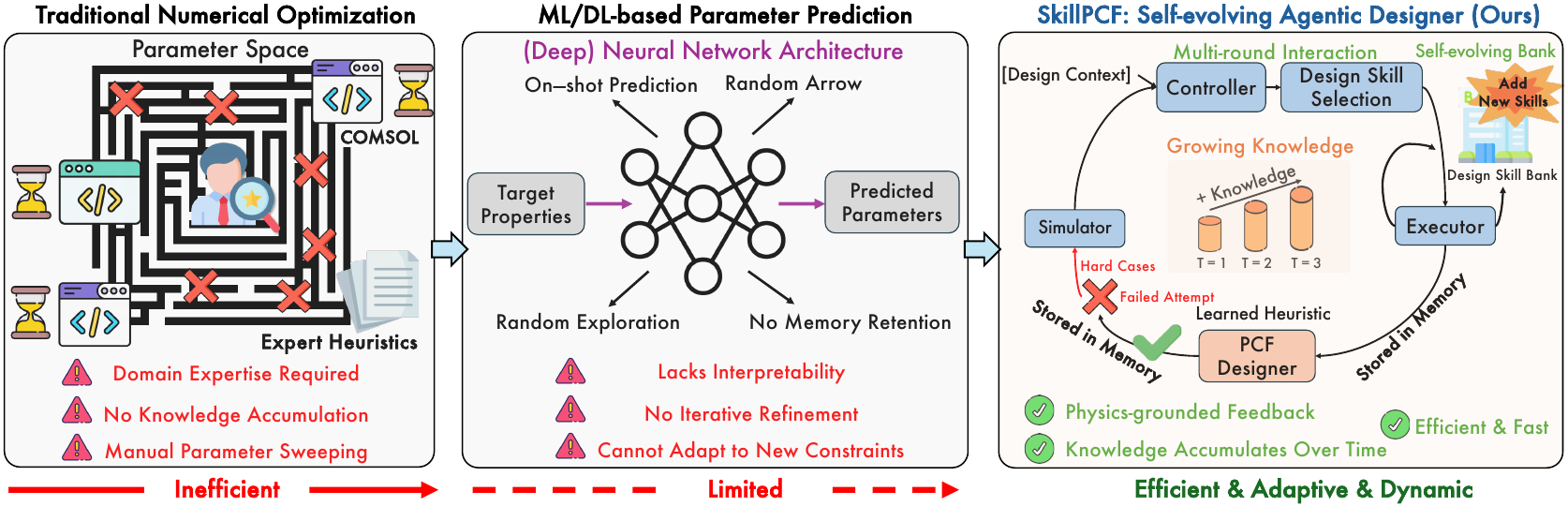}
    \caption{Comparison of PCF inverse design paradigms. \textbf{Left}: Traditional numerical optimization requires domain expertise and exhaustive simulation with no knowledge retention. \textbf{Middle}: ML-based approaches enable fast one-shot prediction but lack iterative refinement and interpretability. \textbf{Right}: SkillPCF (ours) treats design as a multi-round interaction with a self-evolving memory agent, accumulating knowledge across trials and refining skills based on physics-grounded feedback.}
    \label{fig:teaser}
    \vspace{-8pt}
\end{figure*}

Classical PCF studies established geometric principles such as endlessly single-mode guidance and cladding-induced mode control~\cite{Birks_1997}. Recent work introduces design acceleration, including solver-based numerical optimization~\cite{gray2024inversedesignwaveguidedispersion} and machine learning-assisted dispersion-oriented inverse design~\cite{chen2023collaborative, wang2025towards}. These two paradigms are illustrated in Fig.~\ref{fig:teaser} (left and middle): traditional numerical optimization relies on expert heuristics and exhaustive parameter sweeping, while ML-based approaches enable fast one-shot prediction. However, both paradigms treat each design run as independent. The former requires extensive domain expertise with no knowledge retention, and the latter lacks interpretability and iterative refinement. In realistic engineering sessions, designers repeatedly explore nearby regimes; what failed and why, or what worked under which constraints, is itself useful information that should be retained and reused.

This memory gap points to an opportunity: what if a PCF design system could learn to remember what works, forget what does not, and continuously refine its memory strategy from simulation feedback? To our knowledge, no prior work has equipped PCF inverse design with such adaptive memory capabilities. A recent work~\cite{chen2025pcf} explored using LLMs to provide structural understanding and design recommendations, but this approach remains limited to single-turn suggestion generation without iterative memory accumulation or policy learning. Beyond photonics, however, the broader LLM-agent community has made substantial progress on precisely this challenge: how to equip agents with memory systems that store, retrieve, and update information over long horizons~\cite{schick2023toolformer,yao2022react,park2023generative,packer2024memgptllmsoperatingsystems}. Recent work further suggests that memory operations themselves, such as when to insert, update, or delete entries, can be treated as learnable policies rather than fixed heuristics~\cite{shinn2023reflexion,xu2025mem,du2025rethinking}, and that self-evolving agents can discover new memory skills through reinforcement learning~\cite{zhang2026memskill,xia2026skillrl,lu2026choosing}. These developments motivate a question for PCF design: \emph{\textbf{Can PCF inverse design benefit from a memory policy that is learned, domain-conditioned, and continuously refined from simulation outcomes?}}

We argue so. Then we formulate PCF inverse design as a sequence of simulator-guided interactions, where the agent must decide not only what geometry to try next but also what to store, revise, or discard from prior trials. This perspective highlights two practical properties of the setting. First, electromagnetic solvers return deterministic metrics and constraint checks, yielding concrete feedback signals that can be used to study adaptive decision-making~\cite{wen2025reinforcement}. Second, the long-horizon nature of iterative design suggests organizing decisions at multiple levels of abstraction, a viewpoint closely related to hierarchical control~\cite{nachum2018data,dietterich2000hierarchical}.

Building on this perspective, we propose SkillPCF, a memory-centric agent framework for PCF inverse design (Figure~\ref{fig:teaser}, right). SkillPCF formulates design as a multi-round interaction process: a controller selects domain-specific skills, an executor applies them to update a structured memory bank, and a designer refines the skill bank from hard cases, forming a closed loop between physics simulation and memory policy learning. The framework integrates three core components: (i) a physics-guided skill bank for PCF-specific memory operations, (ii) a learnable skill selection controller, and (iii) simulation-grounded rewards within a simulator-in-the-loop training paradigm. To our knowledge, this is the first work to explicitly study a self-evolving LLM-agent memory policy for PCF inverse design. Our contributions are summarized as follows:
\begin{itemize}
\item We \textbf{first} rethink PCF inverse design as a memory-policy learning problem, where reusable design operations are treated as explicit actions optimized via \textbf{deterministic} simulator-grounded feedback.

\item We propose SkillPCF, a closed-loop memory-centric framework integrating a physics-guided skill bank, a learnable skill-selection controller, and a designer module for skill evolution from hard cases.

\item We construct a real-world dataset comprising expert interaction traces, evaluation queries, and multimodal artifacts spanning dispersion engineering, loss optimization, and multi-objective design.

\item We demonstrate through experiments that SkillPCF improves design success and memory effectiveness under low simulation budgets, and that both physics-guided skills and skill evolution are critical.
\end{itemize}

\section{Related Work}
\label{sec:related}

\paragraph{PCF Inverse Design.}
Foundational studies on PCFs established the waveguiding mechanisms and geometric controllability underlying modern PCF design \cite{Russell_2003,Knight_2003,Knight_1996,Birks_1997,chen2022cost}. PCF inverse design is commonly formulated as parameter optimization under highly nonlinear structure--response coupling \cite{gray2024inversedesignwaveguidedispersion}. Traditional methods rely on repeated electromagnetic simulations over large parameter spaces. To improve efficiency, recent work introduces surrogate models and differentiable optimization pipelines to approximate structure--property mappings \cite{wang2025towards,gray2024inversedesignwaveguidedispersion,chen2023collaborative,ren2025unbalanced}. However, these approaches largely operate as parameter-level optimizers or one-shot predictors, without retaining reusable design experience across iterative trials. Rather than replacing physics solvers or optimization routines, we introduce a memory-centric layer that captures transferable design knowledge from multi-round interaction, enabling iterative refinement across PCF design sessions.

\paragraph{Self-evolving LLM Agents.}
Beyond photonics, the LLM-agent community has made substantial progress on memory systems for long-horizon tasks. Tool-augmented agents~\cite{schick2023toolformer,yao2022react} and generative agent simulations~\cite{park2023generative} showed that performance depends critically on how systems store and retrieve interaction history. MemGPT introduced OS-inspired memory hierarchies~\cite{packer2024memgptllmsoperatingsystems}, while Reflexion demonstrated gains from explicit verbal feedback loops~\cite{shinn2023reflexion}. A key insight from recent work is that memory operations, such as when to insert, update, or delete, can be treated as learnable policies rather than fixed heuristics~\cite{xu2025mem,du2025rethinking}, enabling agents to self-evolve their memory skills through reinforcement learning~\cite{zhang2026memskill,xia2026skillrl,lu2026choosing}. This evolution is grounded by verifiable rewards from deterministic environment feedback~\cite{wen2025reinforcement} and structured through hierarchical control~\cite{nachum2018data,dietterich2000hierarchical}. We bring this self-evolving agent paradigm to PCF inverse design, coupling learnable memory skills with physics-grounded simulation feedback.

\section{Method}
\label{sec:method}

\paragraph{Problem Formulation.}
Formally, PCF inverse design aims to find geometric parameters $p\in\mathcal{P}$ such that simulator outputs $f_{\mathrm{sim}}(p)$ jointly satisfy target optical specifications $y^{\star}$ (like dispersion, confinement loss, and effective index) under practical structural constraints. The objective is to maximize target satisfaction while minimizing expensive electromagnetic simulation calls. This is difficult because these targets are coupled and each candidate evaluation requires costly physics simulation.

\begin{figure*}[tbh]
    \centering
    \includegraphics[width=.95\textwidth]{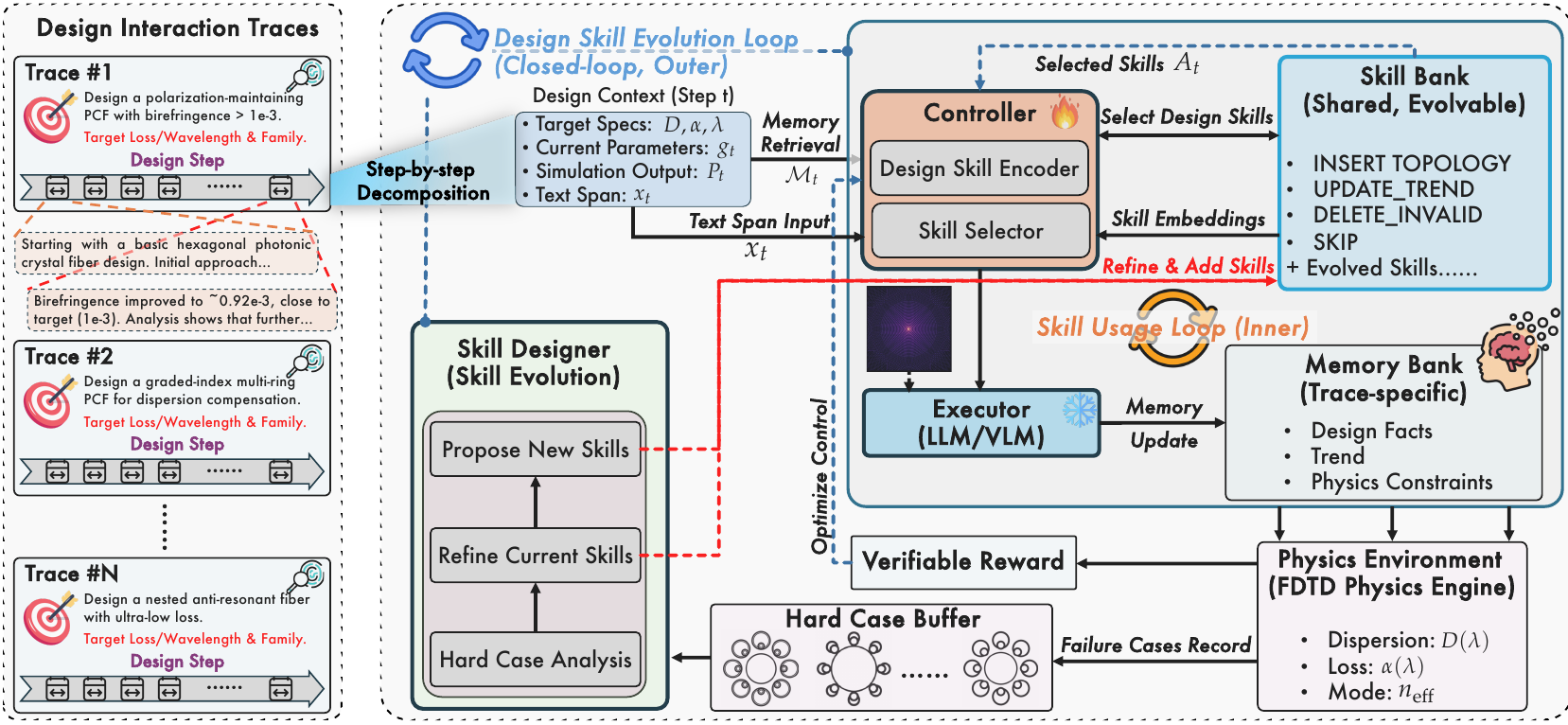}
    \caption{SkillPCF framework overview. Design traces are decomposed into step-wise contexts. At each step $t$, the Controller encodes the design state (target specs, parameters, simulation outputs, and text span) together with retrieved memories, and selects Top-K skills from the Skill Bank. The Executor applies them to update the trace-specific Memory Bank. A Physics Environment supplies simulation-grounded rewards based on dispersion, loss, and modal properties. The Skill Designer leverages hard cases from a failure buffer to refine and expand the skill bank. SkillPCF operates via an inner skill-usage loop (orange) and an outer skill-evolution loop (blue).}
    \label{fig:framework}
    \vspace{-8pt}
\end{figure*}

We propose SkillPCF (Figure~\ref{fig:framework}), a framework that optimizes PCF design through two coupled processes: design skill usage and skill evolution. In the inner loop, a controller selects skills from a shared skill bank conditioned on the current design context, and an executor applies them to update trace-specific memory. In the outer loop, a designer refines existing skills and introduces new ones based on challenging cases encountered during training, enabling continual adaptation of the skill space. To separate trace-level knowledge from reusable memory operations, SkillPCF maintains two distinct stores: a trace-specific memory bank $\mathcal{M}$ and a shared skill bank $\mathcal{S}$. The former accumulates design knowledge for each trajectory, while the latter encodes reusable memory management skills across traces. At outer cycle $e$, we denote the active shared bank by $\mathcal{S}^{(e)}$ (fixed during the inner loop of that cycle). An electromagnetic simulator~\cite{Oskooi2010Meep} provides physics-grounded rewards, forming a simulator-in-the-loop training paradigm. This decoupling of within-episode execution and cross-episode evolution enables stable policy learning while allowing structural adaptation of the skill space.

\subsection{Physics-Guided Skill Bank}
\label{sec:skillbank}

A memory skill defines a reusable, physics-grounded operation for updating design memory within the PCF inverse design loop. Each skill $s \in \mathcal{S}$ consists of (i) a short description used by the controller for selection, and (ii) a structured instruction template executed to produce memory edits. Because PCF inverse design is tightly coupled with electromagnetic simulation, each trajectory span corresponds to a geometry decision and its MEEP-derived physical outcome. Memory therefore stores unit-aware numerical evidence rather than abstract textual summaries. We record both trace-specific outcomes and cross-trace parameter--property relations to preserve physically consistent long-horizon optimization. The skill bank is initialized with four PCF-specific primitives: \textsc{InsertTopologyFeature}, \textsc{UpdatePerformanceTrend}, \textsc{DeleteInvalidAssumption}, and \textsc{Skip}. This minimal set supports stable early learning while allowing subsequent specialization. The controller applies these skills at the span level, and the designer evolves the bank throughout training (see Sections~\ref{sec:learning} and~\ref{sec:designer}); full definitions are provided in Appendix~\ref{app:skills}.

\subsection{Learning Memory Skills for Fine-grained Design}
\label{sec:learning}

\paragraph{Controller: Design Skill Selection.}
\label{sec:controller}
To support fine-grained PCF design, we use a controller that selects a small set of memory skills at each span. We split each trajectory into contiguous spans and process them sequentially, where each span contains the current design action and its simulator context. Let $\mathcal{M}_t$ denote the current trace memory bank and $M_t=\operatorname{Retrieve}(\mathcal{M}_t,x_t)$ the retrieved subset used at step $t$. The controller contains a state encoder $f_{\mathrm{ctx}}$ and a skill encoder $f_{\mathrm{skill}}$. For span $x_t$ and retrieved memory $M_t$, the controller computes:
\begin{equation}
\resizebox{\columnwidth}{!}{$
h_t = f_{\mathrm{ctx}}(x_t, M_t),\ f_{\mathrm{ctx}}(\cdot){=}\operatorname{Embedding}_\mathrm{ctx}(\cdot)\oplus\operatorname{Embedding}_\mathrm{memory}(\cdot).
$}
\end{equation}
where $f_{\mathrm{ctx}}$ concatenates a span embedding with a fused memory embedding (mean or similarity-weighted pooling). For each skill $s_i \in \mathcal{S}^{(e)}$ in the active bank, we compute a skill embedding from its description:
\begin{equation}
\resizebox{\columnwidth}{!}{$
u_i = f_{\mathrm{skill}}(\operatorname{Description}(s_i)),\ f_{\mathrm{skill}}(\cdot){=}\operatorname{Embedding}_\mathrm{skill}(\cdot).
$}
\end{equation}
We use the same embedding model for $f_{\mathrm{ctx}}$ and $f_{\mathrm{skill}}$, so PCF context states and skill descriptions are aligned in one representation space.

\paragraph{Compatibility with Evolving Skill Bank.} During training, the designer may add, remove, or rewrite skills, so the action space changes over time. A fixed-size policy head would break under this evolution. Instead, we score each skill by matching state and skill embeddings as below:
$$z_{t,i} = h_t^\top u_i, \quad p_\theta(i \mid h_t) = \operatorname{Softmax}(z_t)_i,$$
where $z_t \in \mathbb{R}^{|\mathcal{S}^{(e)}|}$ adapts to the current bank size.

\paragraph{Top-K Skill Selection.} Given $p_\theta(i \mid h_t)$, the controller selects an ordered Top-K set $A_t=(a_{t,1},\ldots,a_{t,K})$ without replacement via Gumbel-Top-K sampling~\cite{kool2019stochastic} and passes only these skills to the executor. This is important in PCF traces, where one span may require compound updates (e.g., inserting a new parameter--property fact while deleting an invalid assumption).

\paragraph{Executor: Skill-Conditioned Memory Extraction.}
\label{sec:executor}
Given selected skills $A_t$, executor constructs memory edits by conditioning on (i) the current span $x_t$, (ii) retrieved memory $M_t$, (iii) selected skills $A_t$, and (iv) visual evidence $V_t$. The edits are structured and applied to the trace-specific memory bank:
\begin{equation}
\mathcal{M}_{t+1}=\operatorname{Apply}\big(\mathcal{M}_t,\operatorname{Executor}(x_t,M_t,A_t,V_t)\big).
\end{equation}
Executing multiple skills in one call reduces repeated per-span processing. We optimize the controller with PPO~\cite{Schulman2017Proximal}: for each trajectory, the controller produces a Top-K sequence, the executor builds memory incrementally, and reward is computed from downstream memory-dependent performance. Because actions are ordered and sampled without replacement, the policy probability is:
\begin{equation}
\pi_\theta(A_t \mid h_t) = \prod_{j=1}^{K} \frac{p_\theta(a_{t,j} \mid h_t)}{1 - \sum_{\ell < j} p_\theta(a_{t,\ell} \mid h_t)},
\end{equation}
which reduces to the usual single-action case when $K=1$.

\paragraph{Delayed Reward Redistribution.} In PCF inverse design, whether an early memory decision is useful is often known only after later geometry trials and final verification, which creates sparse credit assignment. In our implementation, the terminal reward is the episode-level QA performance, denoted by $R_{\mathrm{final}}$. We redistribute this delayed signal over spans with exponential decay and keep an explicit terminal portion, the process can be formulated as:
\begin{equation}
\resizebox{\columnwidth}{!}{$
\tilde{r}_t = (1-\beta)\,R_{\mathrm{final}}\,\frac{\gamma^{T-t}}{\sum_{k=1}^{T}\gamma^{T-k}} + \beta\,\mathbf{1}[t{=}T]R_{\mathrm{final}},
$}
\end{equation}
where $\gamma\in(0,1)$ is the redistribution decay and $\beta\in[0,1]$ is the direct terminal ratio. The complete step reward is
\begin{equation}
r_t = r_{\mathrm{proc},t} + \tilde{r}_t,
\end{equation}
where $r_{\mathrm{proc},t}$ is process-level shaping (including memory-construction quality and physics-grounded checks). This form matches the training pipeline and improves long-horizon credit assignment.

\subsection{Feedback-driven Design Skill Evolution}
\label{sec:designer}

Beyond fixed-skill selection, SkillPCF evolves its skill space through a designer module informed by hard PCF cases. At outer cycle $e$, we build a failure buffer $\mathcal{B}^{(e)}$ from low-performing or physically invalid trajectories, where each item contains target specs, predicted design, retrieved memory, and simulator feedback. To improve update diversity, cases in $\mathcal{B}^{(e)}$ are clustered by structural regime and optical-property failure type, and representative samples are selected by difficulty. The designer then performs diagnosis (missing/misaligned operations) followed by refinement (edit existing skills or add structure-aware skills):
\begin{equation}
\hat{\mathcal{S}}^{(e+1)}=\operatorname{Designer}\big(\mathcal{S}^{(e)},\mathcal{B}^{(e)}\big).
\end{equation}
Acceptance, rollback, and post-update exploration scheduling are handled in the closed-loop objective below.

\subsection{Closed-Loop Optimization}
\label{sec:training}
We formulate training as an alternating optimization between controller $\theta$ and the evolving skill bank $\mathcal{S}$. At outer cycle $e$, with fixed skill bank $\mathcal{S}^{(e)}$, the controller is optimized on trajectories by maximizing cumulative step rewards:
\begin{equation}
\begin{aligned}
\theta^{(e+1)} &= \arg\max_{\theta} J\big(\theta;\mathcal{S}^{(e)}\big), \\
J\big(\theta;\mathcal{S}^{(e)}\big) &= \mathbb{E}_{\tau\sim\pi_{\theta}(\cdot\mid\mathcal{S}^{(e)})}\!\left[\textstyle\sum_{t=1}^{T} r_t\right]\!.
\end{aligned}
\end{equation}
From these rollouts, we mine hard cases (failed or low-margin designs) into a buffer $\mathcal{B}^{(e)}$, then let the designer propose an updated bank:
\begin{equation}
\hat{\mathcal{S}}^{(e+1)}=\operatorname{Designer}\big(\mathcal{S}^{(e)},\mathcal{B}^{(e)}\big).
\end{equation}
To prevent drift from physically useful skills, we use acceptance with rollback:
\begin{equation}
\mathcal{S}^{(e+1)}=
\begin{cases}
\hat{\mathcal{S}}^{(e+1)}, & \Delta J_{\mathrm{val}}\ge 0,\\
\mathcal{S}^{(e)}, & \text{otherwise},
\end{cases}
\end{equation}
\begin{equation}
\Delta J_{\mathrm{val}}{=}J_{\mathrm{val}}\big(\theta^{(e+1)},\hat{\mathcal{S}}^{(e+1)}\big){-}J_{\mathrm{val}}\big(\theta^{(e+1)},\mathcal{S}^{(e)}\big).
\end{equation}
After accepted updates, exploration is temporarily biased toward newly introduced design skills, matching the new-action bias mechanism used in training. This closed loop improves both span-level skill usage and cross-episode skill quality under PCF simulator-integrated physics feedback.

\section{Real-World Expert-Annotated Dataset}
\label{sec:dataset}

To enable evaluation of memory-augmented agents for PCF inverse design, we construct a real-world benchmark dataset consisting of four complementary components based on real-world design trajectories and expert insights.

\begin{figure}[tbh]
    \centering
    \includegraphics[width=\columnwidth]{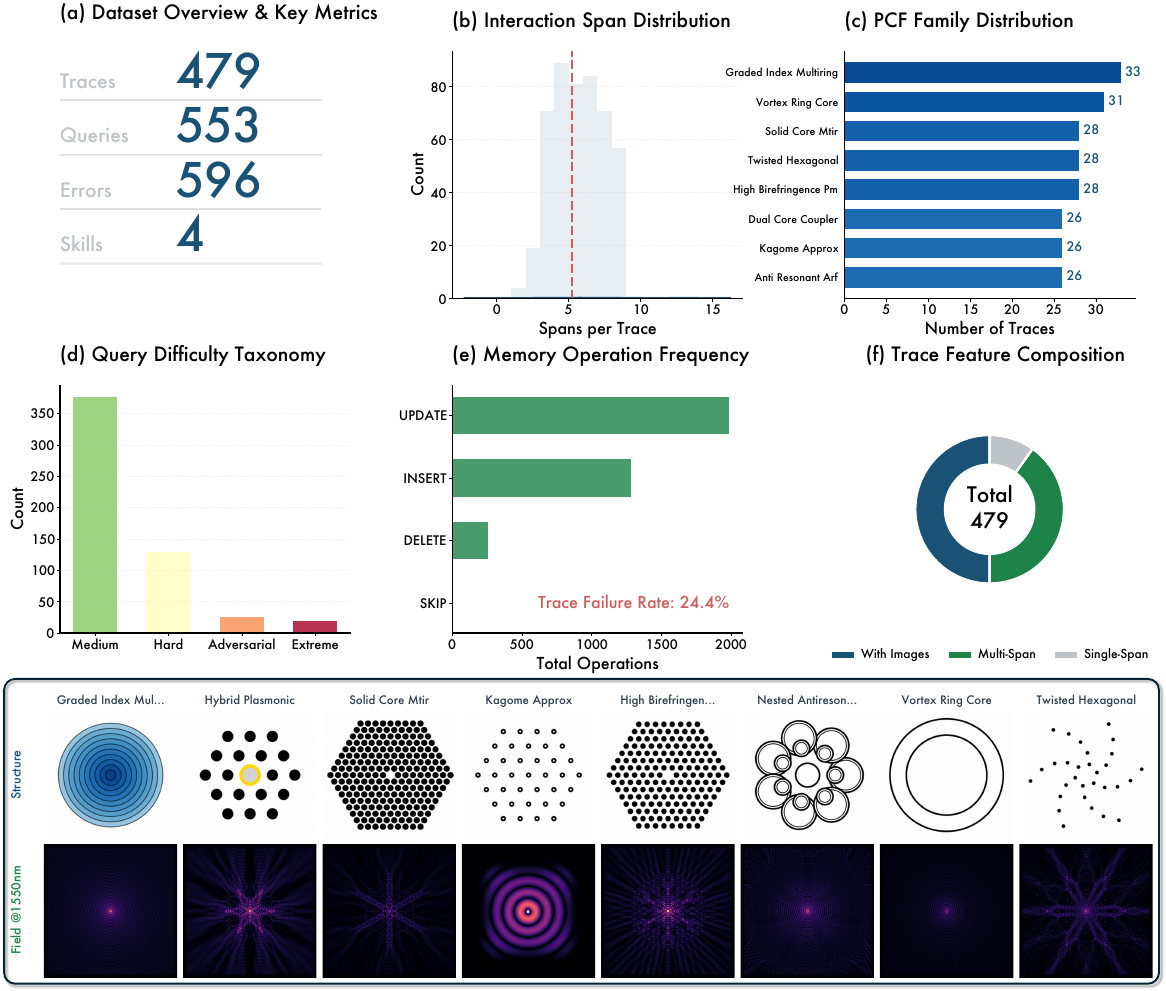}
    \caption{PCFSkill dataset, which contains 479 interaction traces across 8 PCF families, comprising 2,507 spans (5.23 per trace; ~393K tokens) with a 75.6\% design success rate, plus 553 memory-dependent evaluation queries and 596 error logs for skill evolution.}
    \label{fig:dataset}
    \vspace{-10pt}
\end{figure}

\paragraph{Interaction Traces.} We collect 479 expert design trajectories spanning 8 PCF families (e.g., solid-core hexagonal, high-birefringence PM, hollow-core PBG, Kagome, anti-resonant ARF), as shown in Figure~\ref{fig:dataset}. Each trajectory is decomposed into spans (2,507 total, average 5.23 spans per trace), where each span contains a text chunk describing the design action, MEEP simulation code, and the simulation outcome. The dataset comprises approximately 393K tokens with 75.6\% of traces achieving their design goals.

\paragraph{Evaluation Queries.} We construct 553 memory-dependent queries with expert-annotated ground truth. Query types include design reasoning, trend prediction, visual-text fusion, parameter adjustment, multi-hop reasoning, cross-trace comparison, hallucination detection, and failure analysis. Difficulty distribution: medium (68\%), hard (24\%), extreme (4\%), adversarial (5\%). Each query requires 3.49 memory keys on average.

\paragraph{Error Logs.} We collect 596 failure cases for skill evolution, categorized by error type: wrong trend (34\%), missing constraint (31\%), outdated knowledge (18\%), and hallucination (17\%). These logs enable the Designer to identify failure patterns and propose skill refinements.
% Table: Llama4 with/without Visual Field Intervention (side-by-side layout, matching Table 3 style)
\begin{table*}[tbh]
\centering
\caption{Performance of Llama4-Scout~\cite{touvron2023llama} under \emph{Without Visual Field} and \emph{With Visual Field}. Best: bold text with yellow shading; second-best: blue shading.}
\vspace{-6pt}
\label{tab:llama4_comparison}
\resizebox{\textwidth}{!}{%
\begin{tabular}{l|ccccccccc|ccccccccc}
\toprule
& \multicolumn{9}{c|}{\textbf{Without Visual Field}} & \multicolumn{9}{c}{\textbf{With Visual Field}} \\
\cmidrule(lr){2-10} \cmidrule(lr){11-19}
\textbf{Method} & \textbf{Human}$\uparrow$ & \textbf{F1}$\uparrow$ & \textbf{Judge}$\uparrow$ & \textbf{Design}$\uparrow$ & \textbf{Param}$\uparrow$ & \textbf{Trend}$\uparrow$ & \textbf{Succ.}$\uparrow$ & \textbf{Qual.}$\uparrow$ & \textbf{Phys.}$\uparrow$ & \textbf{Human}$\uparrow$ & \textbf{F1}$\uparrow$ & \textbf{Judge}$\uparrow$ & \textbf{Design}$\uparrow$ & \textbf{Param}$\uparrow$ & \textbf{Trend}$\uparrow$ & \textbf{Succ.}$\uparrow$ & \textbf{Qual.}$\uparrow$ & \textbf{Phys.}$\uparrow$ \\
\midrule
RAG          & 7.05 & 17.19 & 7.06 & 20.68 & 60.83 & 69.39 & 32.65 & 5.13 & \second{64.29} & 6.92 & 19.59 & 6.36 & 20.08 & 55.40 & 71.43 & 23.47 & 6.21 & 26.53 \\
CoN          & 6.95 & 14.01 & \second{7.47} & \second{22.97} & 57.10 & 70.41 & 16.33 & 3.90 & \second{64.29} & 6.88 & 16.91 & \second{6.59} & \second{21.32} & 52.89 & 68.37 & 19.90 & 4.60 & 20.41 \\
ReadAgent    & 6.93 & 19.45 & 5.62 & 19.42 & 55.85 & 70.91 & 37.78 & 6.18 & 51.47 & 6.95 & 18.67 & 5.34 & 19.47 & 54.46 & \second{73.47} & \second{40.82} & 5.72 & \second{47.96} \\
MemoryBank   & \second{7.18} & 17.65 & 6.72 & 20.85 & \second{62.15} & 69.39 & 30.61 & 5.58 & 45.92 & 6.98 & 20.40 & 5.71 & 18.57 & 57.69 & 71.43 & 26.02 & 5.73 & 26.53 \\
Mem0         & 5.85 & 18.92 & 5.86 & 19.78 & 42.75 & 70.41 & 3.06 & 1.19 & 6.12  & 5.90 & 18.84 & 5.95 & 18.61 & 43.59 & 66.33 & 6.63 & 4.14 & 5.10 \\
A-MEM        & 6.92 & 20.13 & 6.02 & 19.04 & 58.03 & 68.37 & 36.22 & 7.07 & 38.78 & 6.85 & \second{20.54} & 5.91 & 18.98 & 56.80 & 69.39 & 29.08 & 7.66 & 31.63 \\
LangMem      & 6.85 & \second{20.17} & 5.72 & 19.03 & 58.94 & \second{71.43} & \second{38.27} & 6.69 & 40.82 & 6.82 & 20.24 & 6.05 & 18.20 & \second{58.45} & 70.41 & 36.22 & 7.19 & 32.65 \\
MemoryOS     & 6.80 & 20.08 & 5.87 & 19.68 & 56.48 & 67.35 & 29.08 & \second{12.59} & 37.76 & 6.88 & 20.01 & 5.97 & 19.40 & 52.20 & 68.37 & 31.63 & \second{14.44} & 44.90 \\
\midrule
\textbf{SkillPCF (Ours)} & \best{8.47} & \best{22.35} & \best{8.02} & \best{23.45} & \best{67.28} & \best{82.86} & \best{60.12} & \best{48.73} & \best{68.92} & \best{8.80} & \best{24.18} & \best{7.22} & \best{23.12} & \best{65.85} & \best{82.99} & \best{61.22} & \best{52.35} & \best{71.85} \\
\bottomrule
\end{tabular}%
}
\vspace{-12pt}
\end{table*}

\paragraph{Initial Skill Bank.} We provide four PCF-specific memory skills with structured templates: \texttt{INSERT TOPOLOGY FEATURE} for capturing parameter-property mappings, \texttt{UPDATE PERFORMANCE TREND} for refining relationships, \texttt{DELETE INVALID ASSUMPTION} for removing incorrect heuristics, and \texttt{SKIP} for recognizing when no update is needed. Memory operation distribution across traces: INSERT (36\%), UPDATE (56\%), DELETE (5\%), SKIP (<1\%). Details are in Appendix~\ref{app:skills}.

\section{Experiments and Results}

\paragraph{Dataset and Baselines.} We compare SkillPCF against baselines from two categories: (a) classical numerical optimization methods and (b) memory-augmented agents. Specifically, for classical numerical optimization, we include Random Search, Surrogate Neural Network (NN) Predictor (4 MLP-layers with BN~\cite{ioffe2015batch} and ReLU), and Nelder-Mead~\cite{lagarias1998convergence}. For memory-augmented baselines, we include RAG (Retrieval-Augmented Generation)~\cite{lewis2020retrieval}, CoN (Chain-of-Note)~\cite{yu2024chain}, ReadAgent~\cite{lee2024human}, MemoryBank~\cite{zhong2024memorybank}, Mem0~\cite{chhikara2025mem0}, A-MEM~\cite{xu2025mem}, LangMem, and MemoryOS~\cite{kang2025memory}. Details are in Appendix~\ref{app:baselines}.

\paragraph{Evaluation Protocol and Metrics.} We adopt a multi-level evaluation protocol assessing design quality, simulator cost, and memory effectiveness under identical backbone configurations and fixed trajectory-level splits. Data are partitioned using family-aware stratification (70/15/15 train/validation/test), and retrieval indices for memory baselines are constructed exclusively from training trajectories. At test time, responses are generated with retrieved memories when applicable. If a response proposes a geometry modification or candidate design, we invoke MEEP for deterministic physical verification. We define one simulation call as one full MEEP verification run and report calls per query using this unified definition across all methods. We report metrics along four families that match the column abbreviations in Tables~\ref{tab:llama4_comparison}--\ref{tab:ablation}: \emph{text-based} (token-overlap \textbf{F1}, LLM judge score \textbf{Judge}); \emph{design reasoning} (concept-coverage \textbf{Design}, parameter accuracy \textbf{Param}, trend agreement \textbf{Trend}); \emph{inverse design} (success rate \textbf{Succ.}, design quality \textbf{Qual.}, physics verification \textbf{Phys.}); and \emph{human expert score} (\textbf{Human}, 0--10). Physics verification and design success serve as primary criteria, with the LLM judge used as a supplementary signal. Formal definitions are provided in Appendix~\ref{app:metrics}.

\paragraph{Implementation Details.} We implement SkillPCF in PyTorch~\cite{paszke2019pytorch} with a PPO-based controller and a simulator-in-the-loop executor. GPT-4o-mini~\cite{hurst2024gpt} serves as the LLM judge, Text-Embedding-3-Small encodes states and operations, and Contriever retrieves memories with depth $k=5$. The skill bank is initialized with four operation types (insert, update, delete, no-op), then expanded by the designer during outer-loop evolution. The controller uses an MLP with hidden size 256 and is optimized with AdamW~\cite{loshchilov2017decoupled} ($1\times10^{-4}$). PPO follows standard settings ($\gamma=0.99$, $\lambda=0.95$, clip $0.2$, entropy $0.01$), using four epochs per update, minibatch size 32, and gradient clipping 0.5. Training runs for 10 outer evolution epochs and 50 inner interaction epochs with batch size 32. LLM API calls are routed through OpenRouter; decoding is deterministic with temperature 0 and maximum length 2048. We use a fixed seed for data split and training reproducibility and provide run scripts/configuration in the artifact package. Controller training runs use an NVIDIA RTX A100-40GB GPU. For efficiency comparison, we report Calls/q as a hardware-agnostic simulator-cost proxy.

\begin{table*}[!t]
\centering
\caption{Performance across executor backbones. Calls/q $\downarrow$ is per-query simulator usage (function evaluations for classical optimizers; verification calls for memory agents). Best/second-best highlighted.}
\vspace{-6pt}
\label{tab:backbone_comparison}
\resizebox{.95\textwidth}{!}{%
\begin{tabular}{l|c|ccccccccc|ccccccccc}
\toprule
& \textbf{Human} & \multicolumn{9}{c|}{\textbf{MiniMax-M2.5}~\cite{li2025minimax}} & \multicolumn{9}{c}{\textbf{Qwen2.5-72B}~\cite{bai2023qwen}} \\
\cmidrule(lr){3-11} \cmidrule(lr){12-20}
\textbf{Method} & \textbf{Score}$\uparrow$ & \textbf{F1} & \textbf{Judge} & \textbf{Design} & \textbf{Param} & \textbf{Trend} & \textbf{Succ.} & \textbf{Qual.} & \textbf{Phys.} & \textbf{Calls/q}$\downarrow$ & \textbf{F1} & \textbf{Judge} & \textbf{Design} & \textbf{Param} & \textbf{Trend} & \textbf{Succ.} & \textbf{Qual.} & \textbf{Phys.} & \textbf{Calls/q}$\downarrow$ \\
\midrule
\multicolumn{20}{l}{\textit{Memory-Augmented Agents}} \\
\midrule
RAG & 7.10 & 22.41 & 6.08 & 19.76 & 55.62 & 68.37 & 42.86 & 6.88 & 57.14 & 1.13$^\dagger$ & 21.77 & 5.87 & 20.20 & 56.08 & 68.37 & 35.71 & 5.54 & 54.08 & 1.13 \\
CoN & 6.98 & 20.12 & 6.21 & \second{20.58} & 54.91 & 66.33 & 40.82 & 7.21 & 55.10 & 1.12$^\dagger$ & 19.15 & 6.03 & \second{21.90} & 56.77 & 65.31 & 35.71 & 6.58 & 54.08 & 1.12 \\
ReadAgent & 6.86 & \second{23.67} & 5.96 & 18.92 & \second{56.45} & 69.39 & 37.76 & 7.06 & 49.98 & 1.13$^\dagger$ & \second{22.34} & 5.58 & 19.04 & 56.23 & 68.37 & 33.16 & 6.50 & 46.94 & 1.13 \\
MemoryBank & \second{7.22} & 22.58 & 6.18 & 19.85 & 55.93 & 70.41 & 46.94 & 9.12 & 56.63 & 1.12$^\dagger$ & 21.05 & 6.11 & 19.46 & 54.83 & 69.39 & 43.67 & 8.66 & 54.08 & 1.12 \\
Mem0 & 5.92 & 21.16 & \second{6.39} & 17.31 & 42.38 & \second{71.43} & 3.06 & 0.98 & 16.33 & \second{1.03} & 22.07 & \second{6.30} & 18.43 & 47.44 & 68.37 & 12.76 & 3.01 & 39.80 & 1.11 \\
A-MEM & 6.88 & 22.98 & 6.02 & 19.11 & 53.51 & 67.35 & 50.00 & 8.22 & 48.98 & 1.15 & 21.21 & 6.00 & 20.79 & \second{57.12} & \second{70.41} & 41.33 & 8.95 & 52.04 & \second{1.10} \\
LangMem & 6.79 & 23.05 & 5.44 & 17.16 & 54.91 & 69.39 & 38.78 & 8.56 & 44.90 & 1.20 & 21.59 & 6.01 & 19.87 & 55.41 & 67.35 & 41.33 & 6.77 & 57.14 & 1.17 \\
MemoryOS & 6.83 & 23.26 & 5.46 & 19.11 & 54.91 & 67.35 & 38.78 & 14.76 & 55.10 & 1.15 & 20.55 & 5.68 & 18.12 & 54.47 & 67.35 & 24.49 & 8.97 & 60.20 & 1.15 \\
\midrule
\textbf{SkillPCF (Ours)} & \best{9.12} & \best{32.85} & \best{6.92} & \best{24.35} & \best{62.24} & \best{84.85} & 82.35 & \second{52.18} & \best{68.45} & \best{1.02} & \best{25.12} & \best{7.95} & \best{29.85} & \best{69.35} & \best{72.45} & 78.92 & \second{48.76} & \best{65.28} & \best{1.02} \\
\midrule
\multicolumn{20}{l}{\textit{Classical Optimization}$^\dagger$ ($\dagger$ They are backbone-independent and do not rely on external LLMs, thereby maintaining consistency across backbone configurations)} \\
\midrule
RandomSearch & 6.05 & 15.00 & 1.60 & 12.90 & 11.50 & 19.40 & \best{92.90} & 7.30 & \second{64.30} & 100.00 & 15.00 & 1.60 & 12.90 & 11.50 & 19.40 & \best{92.90} & 7.30 & \second{64.30} & 100.00 \\
NNPredictor & 4.62 & 13.60 & 1.80 & 12.70 & 15.70 & 19.40 & 0.00 & 0.00 & 63.30 & 100.00 & 13.60 & 1.80 & 12.70 & 15.70 & 19.40 & 0.00 & 0.00 & 63.30 & 100.00 \\
NelderMead & 6.42 & 12.20 & 2.40 & 12.40 & 31.40 & 19.40 & \second{91.80} & \best{81.60} & \second{64.30} & 135.00 & 12.20 & 2.40 & 12.40 & 31.40 & 19.40 & \second{91.80} & \best{81.60} & \second{64.30} & 135.00 \\
\bottomrule
\end{tabular}%
}
\vspace{-10pt}
\end{table*}

\subsection{Main Results}
\label{sec:results}

Tables~\ref{tab:llama4_comparison} and~\ref{tab:backbone_comparison} show that SkillPCF delivers the strongest overall performance among memory-agent methods while retaining low simulation cost. Under Llama4-Scout, SkillPCF ranks first across all reported metrics in both settings (without/with visual field). Cross-backbone results further confirm robustness: SkillPCF remains best on MiniMax-M2.5 and Qwen2.5-72b, and a constant 1.02 calls per query. Other memory-agent baselines operate at similar call budgets (typically 1.10--1.20) but consistently lower Success and Human scores, indicating that gains come from higher-quality memory policy rather than extra simulator usage. Classical optimizers can reach high raw Success under exhaustive search, yet they require 100--135 calls per query and provide markedly weaker reasoning quality, which limits practicality in interactive design workflows.

\subsection{Framework Analysis}

\begin{figure}[tbh]
    \centering
    \includegraphics[width=\columnwidth]{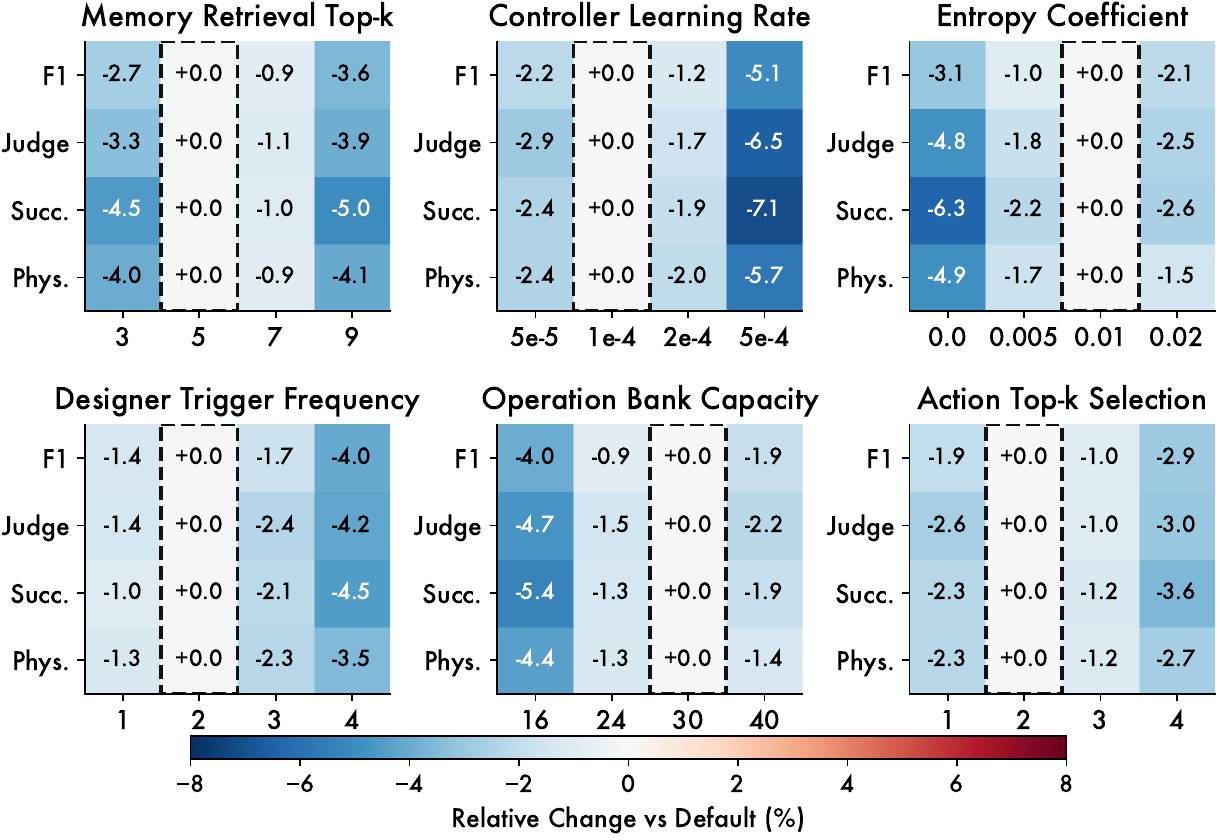}
    \caption{Hyperparameter sensitivity. Cells report relative (\%) changes from default settings across F1, Judge, Success, and Physics metrics; dashed columns denote defaults.}
    \label{fig:hparam_sensitivity}
    \vspace{-10pt}
\end{figure}

\paragraph{Hyperparameter Sensitivity.} Figure~\ref{fig:hparam_sensitivity} shows that SkillPCF remains stable under moderate hyperparameter variation around the default configuration. Across retrieval depth, learning rate, entropy, designer frequency, operation-bank size, and action top-$k$, performance typically varies within $\pm$8\% of baseline. Noticeable degradation occurs only under overly large learning rates or insufficient exploration regularization. The default sits in a broad stable region, indicating hyperparameter robustness.

\paragraph{Ablation Studies.} Table~\ref{tab:ablation} shows that performance gains arise from complementary and non-redundant components. Removing the skill controller leads to the largest Success drop and reduced Human score, highlighting the importance of learned operation selection. Removing physics reward shaping results in the strongest PhysicsV decline, underscoring the role of simulator-grounded rewards in physical validity. Disabling designer evolution or reward redistribution further reduces F1, Success, and Human scores, though to a lesser extent. Visual intervention helps physics-grounded metrics (F1, Succ., Phys.) but slightly lowers LLM-judged Design/Judge, which we attribute to the judge's weaker calibration on mode-field images. We therefore treat physics-grounded metrics as primary.
\begin{table*}[tbh]
\centering
\caption{Ablation on Llama4-Scout (visual). Best/second-best highlighted in yellow/blue.}
\vspace{-6pt}
\label{tab:ablation}
\resizebox{.8\textwidth}{!}{%
\begin{tabular}{lccccccccc}
\toprule
\textbf{Variant} & \textbf{F1} $\uparrow$ & \textbf{Judge} $\uparrow$ & \textbf{Design} $\uparrow$ & \textbf{Param} $\uparrow$ & \textbf{Trend} $\uparrow$ & \textbf{Succ} $\uparrow$ & \textbf{Qual} $\uparrow$ & \textbf{PhysV} $\uparrow$ & \textbf{Human} $\uparrow$ \\
\midrule
SkillPCF (Full) & \best{24.18} & \second{7.22} & \second{23.12} & \second{65.85} & \best{82.99} & \best{61.22} & \best{52.35} & \best{71.85} & \best{8.80} \\
\midrule
\emph{w/o} Designer Evolution & 22.94 & 6.89 & 21.76 & 63.10 & 80.71 & 54.36 & 41.82 & 66.34 & 7.85 \\
\emph{w/o} Physics Reward Shaping & 23.36 & 7.03 & 22.41 & 64.27 & 81.38 & 49.28 & 38.47 & 58.92 & 7.53 \\
\emph{w/o} Adaptive Skill Controller & 21.88 & 6.74 & 20.95 & 61.90 & 79.66 & 47.55 & 35.41 & 61.08 & 7.28 \\
\emph{w/o} New-Action Bias & \second{23.57} & 7.06 & 22.66 & 64.88 & 82.05 & 56.74 & 46.29 & \second{69.14} & \second{8.18} \\
\emph{w/o} Reward Redistribution & 22.81 & 6.93 & 21.89 & 63.52 & 80.97 & 53.11 & 43.38 & 65.27 & 7.67 \\
\emph{w/o} Visual Intervention & 22.35 & \best{8.02} & \best{23.45} & \best{67.28} & \second{82.86} & \second{60.12} & \second{48.73} & 68.92 & 8.05 \\
\bottomrule
\end{tabular}%
}
\vspace{-6pt}
\end{table*}

\begin{figure}[tbh]
    \centering
    \includegraphics[width=\columnwidth]{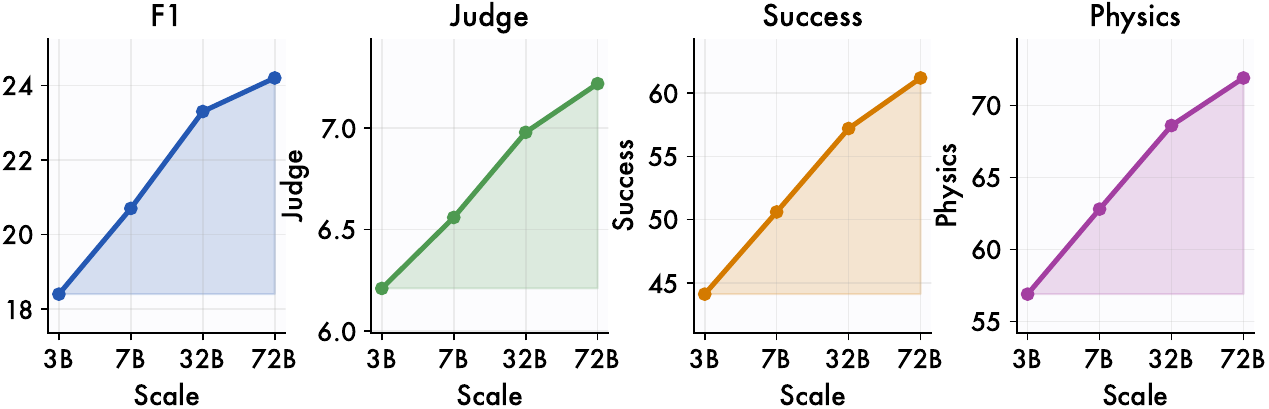}
    \caption{Scaling under Qwen2.5-VL executor (3B/7B/32B/72B).}
    \label{fig:scaling_behavior}
    \vspace{-12pt}
\end{figure}

\paragraph{Scaling Behaviors.} Figure~\ref{fig:scaling_behavior} evaluates executor scaling from Qwen2.5-VL 3B to 72B with the controller and skill bank fixed. All metrics improve monotonically as model size increases, with diminishing gains beyond 32B. The 32B model already attains performance close to 72B, indicating a favorable quality--cost trade-off at mid scale. Most performance gains are therefore captured before the largest model sizes, suggesting efficient scaling behavior.

\begin{figure}[tbh]
    \centering
    \includegraphics[width=\columnwidth]{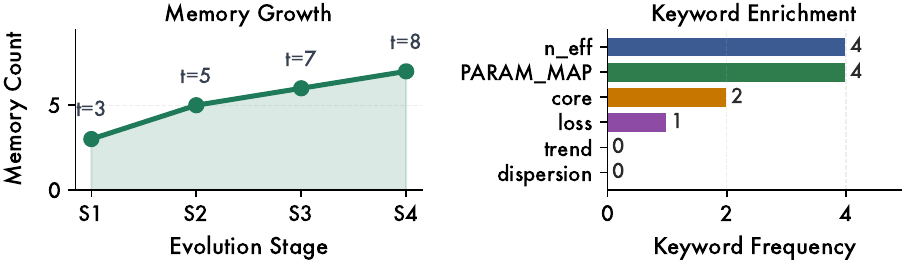}
    \caption{Memory dynamics during skill evolution. Left: memory growth across outer-loop. Right: keyword enrichment.}
    \label{fig:skill_evolution_main}
    \vspace{-8pt}
\end{figure}

\paragraph{Memory Dynamics under Skill Evolution.} Figure~\ref{fig:skill_evolution_main} reports two statistics illustrating how hard-case feedback reshapes memory across outer-loop stages. The left panel shows monotonic growth in memory size at selected checkpoints, while the right panel reflects increased domain relevance through keyword frequency and operation usage. These results suggest that designer-triggered updates improve not only memory quantity but also its physical informativeness. Examples of refined and newly introduced skills are provided in Appendix~\ref{app:skills}.

\subsection{Case Studies}
\label{sec:case_studies}
\begin{figure}[tbh]
    \centering
    \includegraphics[width=\columnwidth]{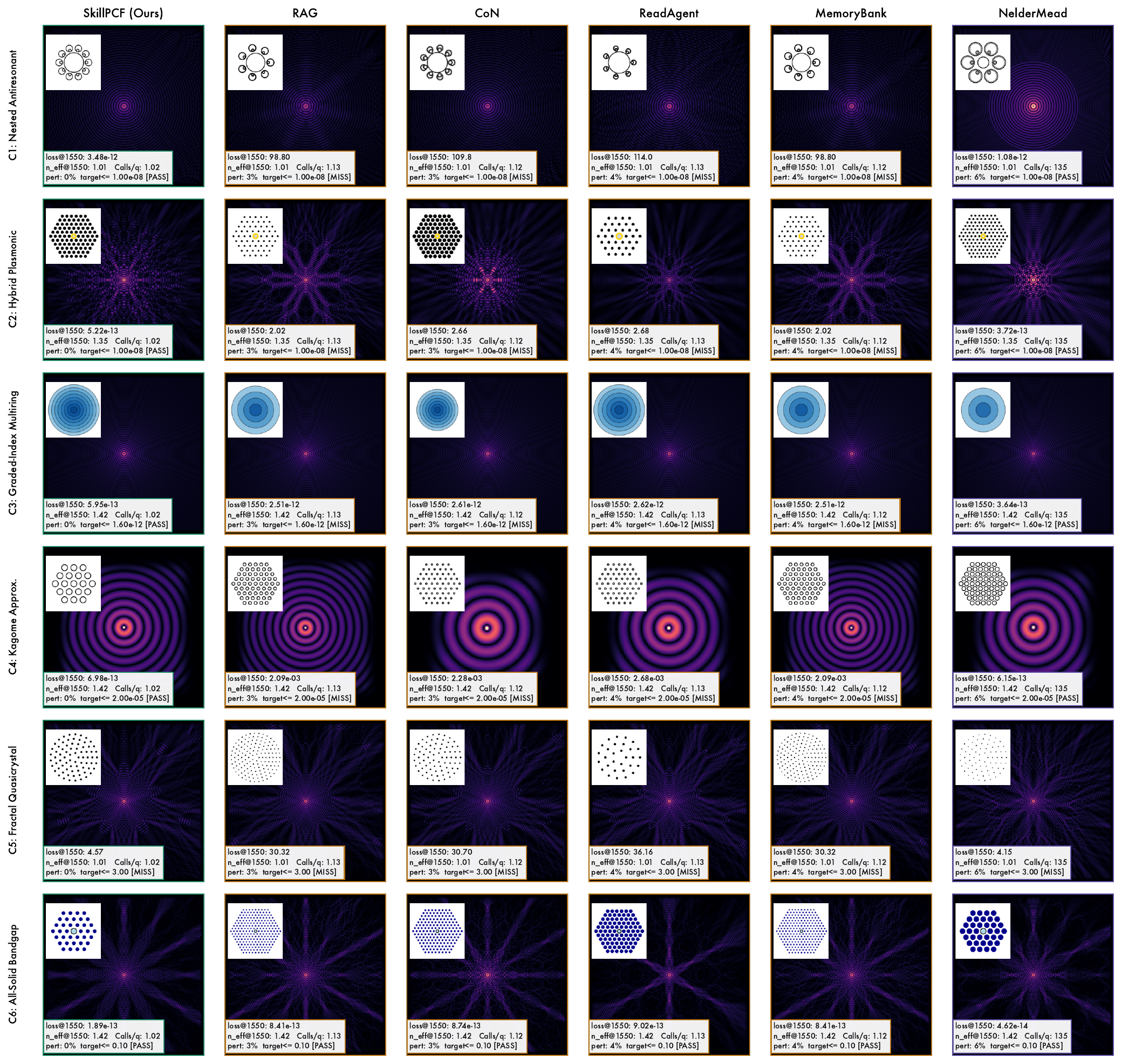}
    \caption{Case study across six PCF families. Each panel: mode field (background), structure (inset), key metrics (loss, $n_{\text{eff}}$, calls/query). \textbf{PASS}/\textbf{FAIL} badges mark target satisfaction. Best viewed zoomed in and in color. Extended studies in Appendix~\ref{app:extended_case_studies}.}
    \label{fig:case_baselines}
    \vspace{-10pt}
\end{figure}

We conduct a case study on six challenging PCF design tasks (Figure~\ref{fig:case_baselines}). SkillPCF satisfies target specifications on five of six families, with simulated mode fields exhibiting clear confinement and losses meeting stringent thresholds (e.g., $\leq 10^{-8}$ dB/km for nested antiresonant); the remaining case (C5, fractal quasicrystal) is missed by 0.93 dB/km and exposes a hard regime where the current skill bank lacks fractal-symmetry priors. In contrast, baselines show sharper sensitivity to difficulty, succeeding only under relaxed targets and failing under stricter constraints such as graded-index multiring at $1.6\times 10^{-12}$ dB/km. Memory-based methods use comparable budgets (1.1--1.3 calls/query); gaps reflect memory quality and skill selection, not simulator usage. Per-family quantitative breakdowns and extended qualitative inspection are provided in Appendix~\ref{app:extended_case_studies}.

\begin{figure}[tbh]
    \centering
    \includegraphics[width=\columnwidth]{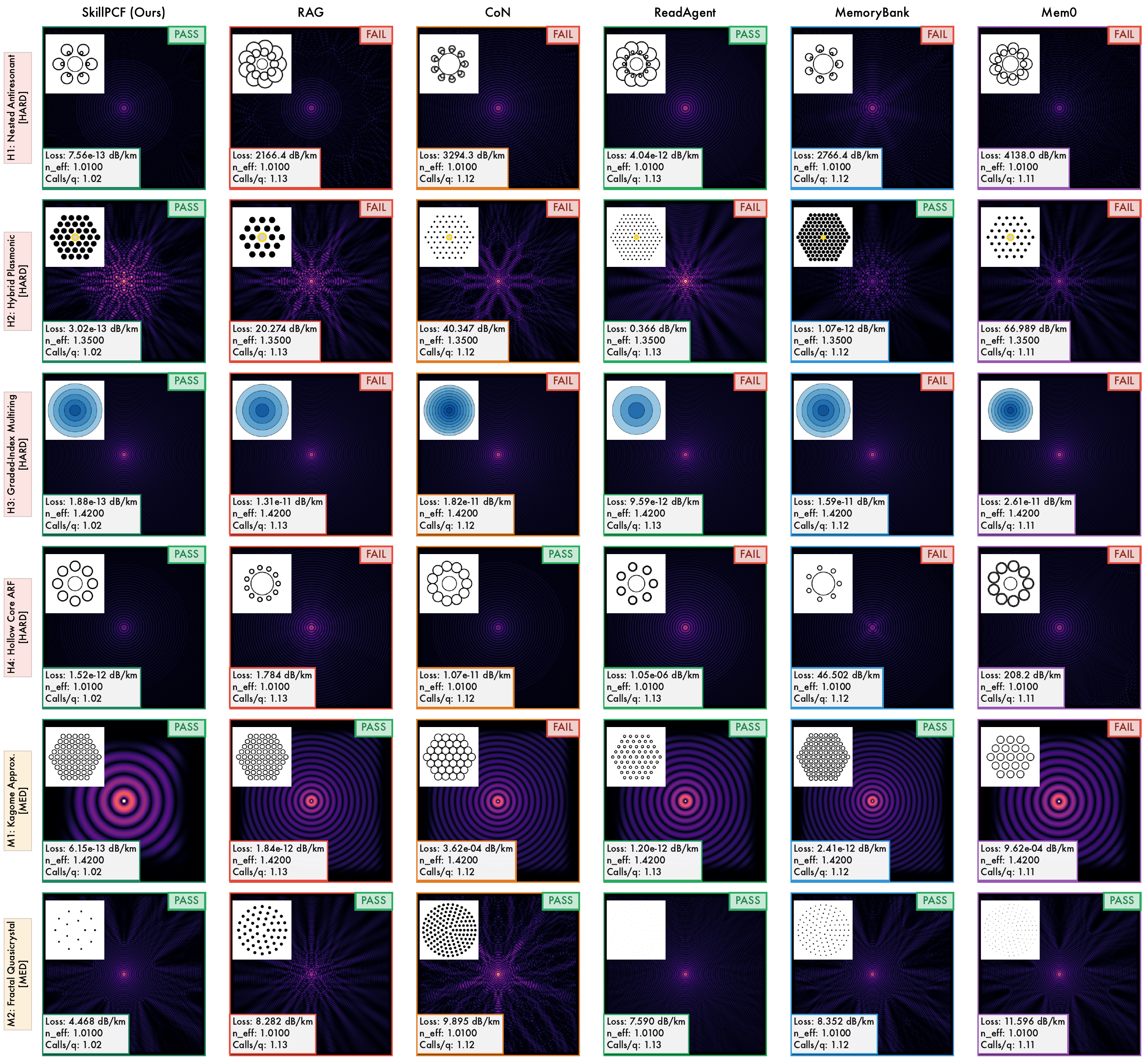}
    \caption{Hard-case stress test with real MEEP simulations across six PCF scenarios. Rows are design cases (H: hard, M: medium); columns are methods. Best viewed zoomed in and in color.}
    \label{fig:hardcase_main}
    \vspace{-10pt}
\end{figure}

\paragraph{Hard-Case Test.} Figure~\ref{fig:hardcase_main} stresses methods on ultra-low-loss targets ($<10^{-8}$ dB/km), zero-dispersion designs, and novel architectures (rows H1--H4) alongside standard telecom cases (M1--M2). Baselines mostly fail on H1--H4 with isolated passes; SkillPCF passes consistently across difficulty levels under matched simulation budgets, indicating that memory accumulation translates into genuine physical-design improvements rather than higher solver effort.

\begin{figure}[!t]
    \centering
    \includegraphics[width=\columnwidth]{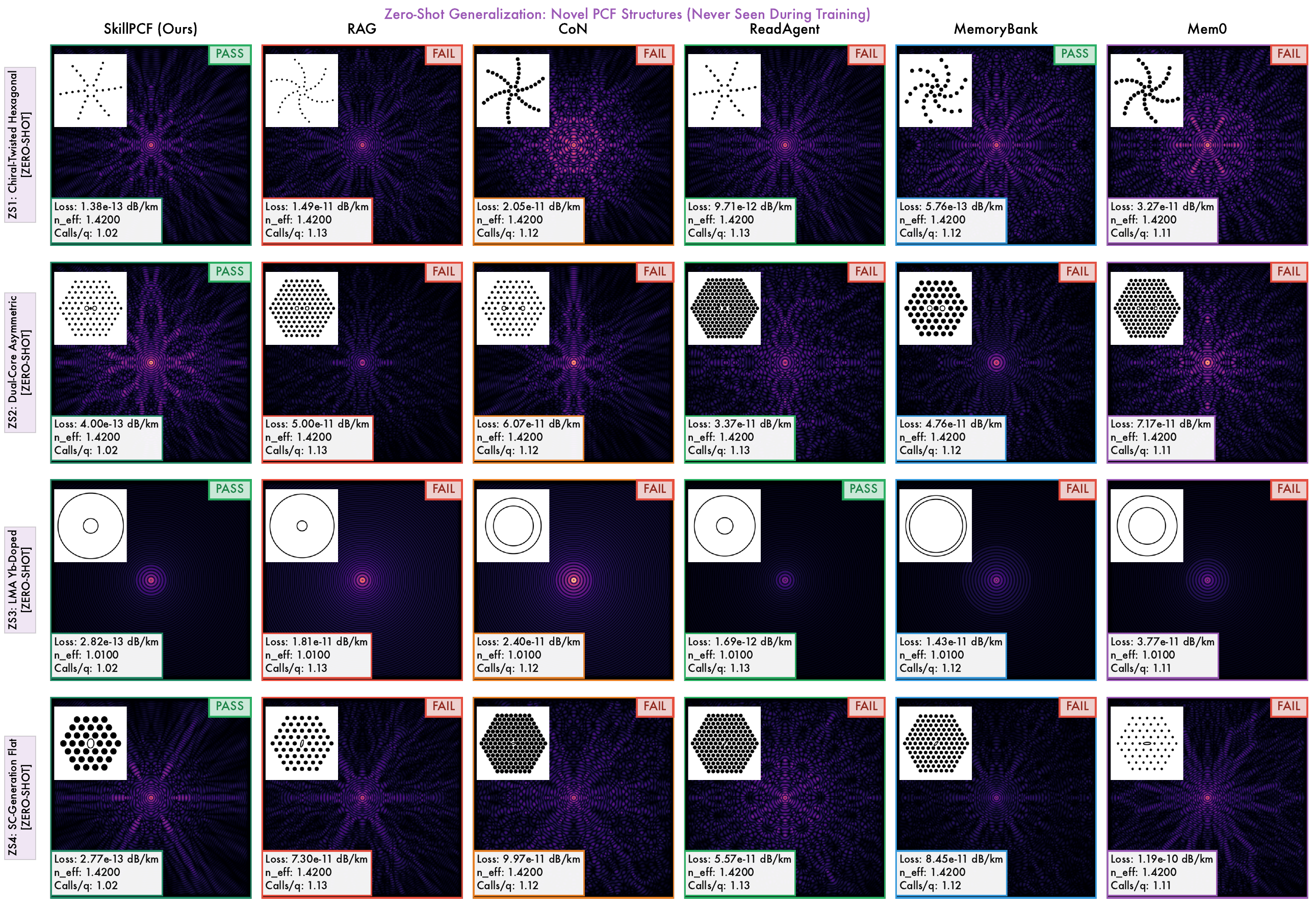}
    \caption{Zero-shot generalization to four novel PCF families never seen in training. Rows: ZS1 chiral-twisted hexagonal; ZS2 dual-core asymmetric coupler; ZS3 LMA Yb-doped; ZS4 SC-generation flat. Best viewed zoomed in and in color.}
    \label{fig:zeroshot_main}
    \vspace{-10pt}
\end{figure}

\paragraph{Zero-Shot Generalization.} Figure~\ref{fig:zeroshot_main} evaluates four \emph{novel} PCF families never seen in training (chiral-twisted hexagonal, dual-core asymmetric coupler, LMA Yb-doped, SC-generation flat). SkillPCF transfers consistently while baselines show only sporadic passes, suggesting that learned skills capture transferable physics relationships such as confinement, dispersion control, and mode engineering, rather than parameter correlations specific to seen families.

\section{Conclusion}
We propose SkillPCF, a memory-centric framework for PCF inverse design that models reusable design experience as a learnable policy. By integrating skill selection, memory updates, and simulator verification, it transforms iterative traces into transferable optimization knowledge. Results show strong target satisfaction with high simulation efficiency, supporting physics-aware design automation.

\clearpage

\section*{Impact Statement}

SkillPCF reduces the compute and human effort of PCF inverse design by reusing physics-grounded design experience across simulation budgets, with downstream relevance to optical-fiber R\&D in telecommunications, sensing, and biomedical imaging, and to other simulator-grounded design domains such as metamaterials and antennas. The principal risk is over-reliance on language-model-suggested geometries, mitigated by deterministic MEEP verification of every candidate and by releasing queries, rubrics, and run scripts. Dependence on hosted foundation models is addressed via fixed seeds, stratified splits, and full configuration disclosure. We do not foresee further societal consequences warranting specific discussion.

\bibliography{main}
\bibliographystyle{icml2026}

%%%%%%%%%%%%%%%%%%%%%%%%%%%%%%%%%%%%%%%%%%%%%%%%%%%%%%%%%%%%%%%%%%%%%%%%%%%%%%%
%%%%%%%%%%%%%%%%%%%%%%%%%%%%%%%%%%%%%%%%%%%%%%%%%%%%%%%%%%%%%%%%%%%%%%%%%%%%%%%
% APPENDIX
%%%%%%%%%%%%%%%%%%%%%%%%%%%%%%%%%%%%%%%%%%%%%%%%%%%%%%%%%%%%%%%%%%%%%%%%%%%%%%%
%%%%%%%%%%%%%%%%%%%%%%%%%%%%%%%%%%%%%%%%%%%%%%%%%%%%%%%%%%%%%%%%%%%%%%%%%%%%%%%
\newpage
\appendix
\onecolumn
\section*{Appendix}
\noindent For readability and fast lookup, we organize the appendix into six blocks with direct hyperlinks:
\begin{itemize}
  \item \hyperref[app:a]{Appendix A: System, Data, and Evaluation Protocols}
  \item \hyperref[app:b]{Appendix B: Additional Case Studies}
  \item \hyperref[app:c]{Appendix C: Implementation Details and Evaluation Metrics}
  \item \hyperref[app:d]{Appendix D: Initial PCF Memory Skills}
  \item \hyperref[app:e]{Appendix E: Skill Evolution Examples During Training}
  \item \hyperref[app:f]{Appendix F: Extended Case Study Analysis}
\end{itemize}

\paragraph{Background.} Photonic crystal fibers sit within a broader trend of applying machine learning to fiber-optic and photonic systems, where data-driven models have improved tasks ranging from wavelength interrogation and interferometric sensor demodulation to spectral reconstruction from sparse measurements and efficient calculation of optical properties for microstructured fibers \cite{chen2022cost,chen2022fabry,chen2022reconstruction,hou2024federated,ren2023unsupervised,ren2024rethink,wang2025optimizing,yuan2022efficient}. These efforts established that learned surrogates can replace or accelerate expensive optical characterization, and the perspective has recently been extended to PCF reverse design through distributed knowledge-fusion frameworks \cite{ren2024distributed}. SkillPCF continues this trajectory but shifts the locus of learning from one-shot prediction toward a memory policy that accumulates and reuses design experience across iterative, simulator-grounded trials.

\section{System, Data, and Evaluation Protocols}
\label{app:a}
\label{app:system_expert}

\subsection{Front-end System Overview}

\begin{figure}[tbh]
    \centering
    \includegraphics[width=\textwidth]{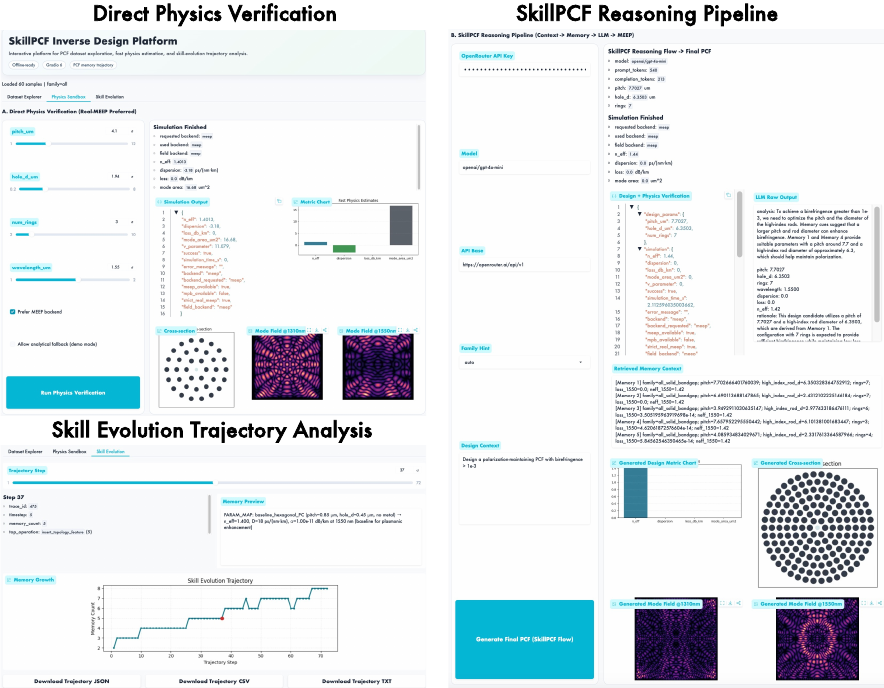}
    \caption{Interactive SkillPCF design platform. Left: coupled physics verification and skill-evolution trajectory analysis, including parameter controls, synchronized structure/field, and round-level skill updates. Right: memory-augmented reasoning pipeline with design-context input, memory retrieval, generation, and verification feedback.}
    \label{fig:system}
\end{figure}

For the ease of user interaction, we design a front-end system for our SkillPCF-based PCF inverse design. Our interactive design platform integrates real-time electromagnetic simulation with memory-augmented design reasoning, as illustrated in Figure~\ref{fig:system}. The system architecture comprises two synchronized panels:

\paragraph{Left panel (Physics + Skill Evolution Analysis).} The left side contains two coordinated modules. The upper module provides direct access to the MEEP simulation backend with interactive parameter controls for all PCF geometric variables (air-hole diameters $d_i$, lattice pitch $\Lambda$, core defect configurations), along with synchronized structure/field visualization and metric feedback (effective index $n_{\text{eff}}$, confinement loss $\alpha$, dispersion $D$). The lower module visualizes skill evolution trajectory analysis across optimization rounds, exposing which memory skills are activated, refined, or deprecated as solver feedback accumulates.

\paragraph{Right panel (Memory-Augmented Reasoning).} Exposes the complete SkillPCF reasoning pipeline. The design-context input module accepts natural language task specifications (e.g., ``design PCF with near-zero dispersion at 1550 nm''). The retrieved-memory display shows relevant prior design knowledge extracted from the skill bank, including parameter-property mappings, constraint records, and successful strategy patterns. The generated output module presents the agent's design recommendations with explicit reasoning traces. Post-generation verification validates proposed designs against target specifications and feeds outcomes back to both memory updates and trajectory analytics.

The platform supports iterative design workflows where users can accept, modify, or reject agent suggestions, with all interactions logged for benchmark construction and skill evolution analysis. All system code, trained models, and benchmark datasets will be made publicly available upon paper acceptance to facilitate reproducibility and community research.

\subsection{Expert Team and Dataset Construction Protocol}

The expert team composition is detailed in Table~\ref{tab:dataset_expert_team}. Dataset construction followed a three-stage protocol: (i) trajectory drafting from realistic inverse-design tasks, (ii) simulation-backed verification with unit and constraint checks, and (iii) independent audit of annotations and query-ground-truth mappings. Ambiguous traces were returned for revision until both physical consistency and textual clarity passed review, yielding a reliable benchmark for memory-dependent reasoning. For evaluation governance and scoring anchors, see Table~\ref{tab:human_eval_process} and Table~\ref{tab:human_scoring_rubric}.

\begin{table}[!ht]
\centering
\footnotesize
\caption{Expert composition used in dataset construction.}
\label{tab:dataset_expert_team}
\begin{tabular}{p{0.27\linewidth}p{0.10\linewidth}p{0.57\linewidth}}
\toprule
\textbf{Role} & \textbf{Count} & \textbf{Background and responsibility} \\
\midrule
PCF design researchers & 4 & PhD-level photonics background (4--10 years); authored design trajectories, defined target specifications, and validated physical plausibility. \\
Simulation specialists & 2 & Electromagnetic simulation experience with MEEP/MPB (3--8 years); audited solver settings, unit consistency, and convergence checks. \\
Annotation auditors & 2 & Research staff with optical-device data curation experience (2--5 years); normalized span annotations, query labels, and cross-check logs. \\
\bottomrule
\end{tabular}
\end{table}

\subsection{Human-Evaluation Governance Summary}

To avoid fragmented reporting, we consolidate full human-evaluation protocol details (workflow, fairness controls, and scoring rubric) with the metric definitions in Appendix~\ref{app:c}. This keeps all quantitative evaluation standards in one place while preserving dataset-construction context.

\section{Additional Case Studies}
\label{app:b}

This section provides supplementary qualitative evidence complementary to the main case-study analysis.

\subsection{Classical-Budget Comparison}

\begin{figure}[tbh]
    \centering
    \includegraphics[width=\textwidth]{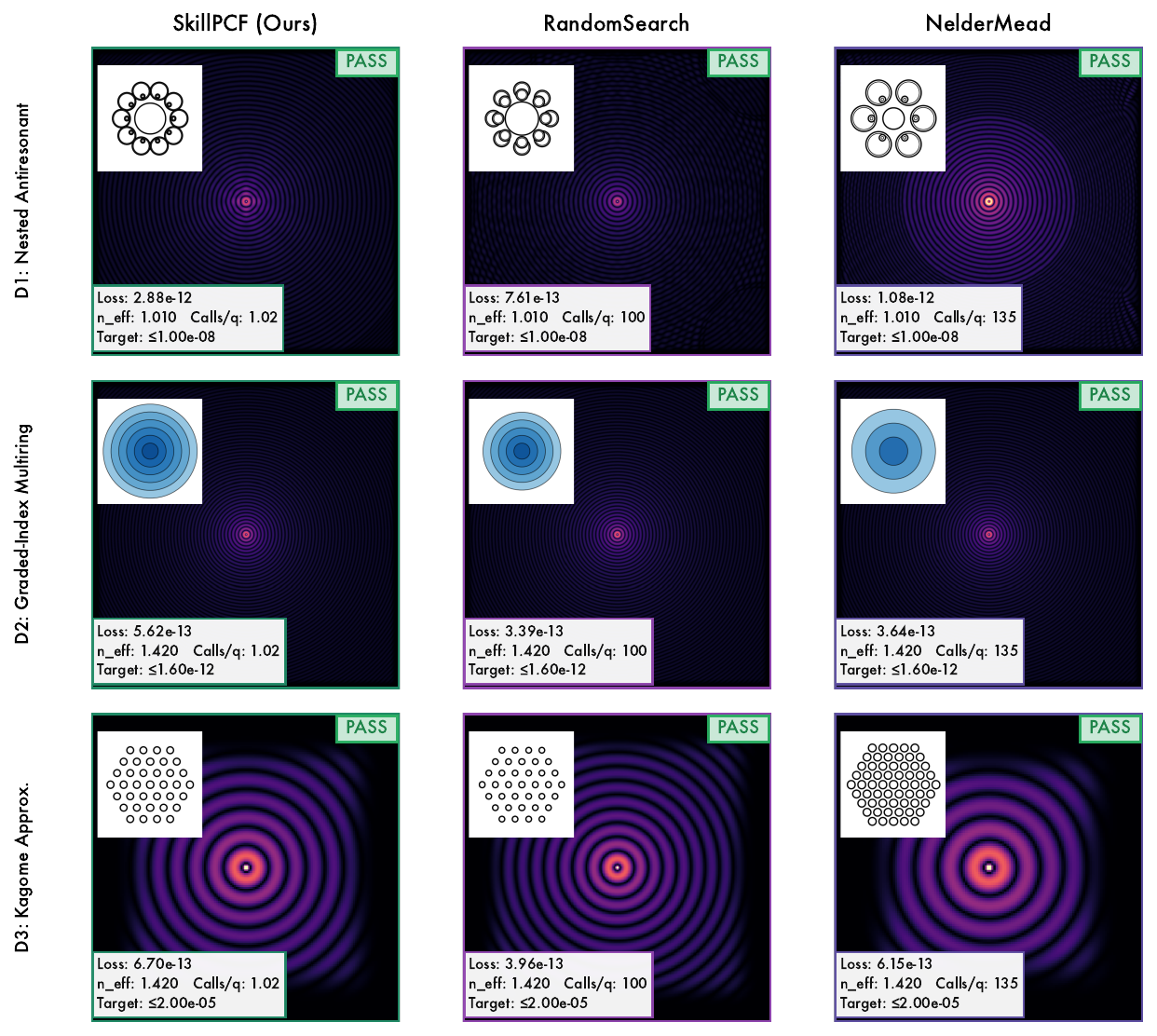}
    \caption{Efficiency comparison between SkillPCF and classical optimization methods.}
    \label{fig:case_classical}
\end{figure}

Figure~\ref{fig:case_classical} further compares SkillPCF against classical optimization baselines. While RandomSearch and NelderMead achieve competitive success rates (92.9\% and 91.8\% respectively), they require approximately 100--135 simulation calls per query, two orders of magnitude more than SkillPCF. This efficiency gap is critical for practical PCF design, where each electromagnetic simulation may take minutes to hours depending on structure complexity. The results demonstrate that SkillPCF offers a favorable trade-off: near-classical design quality at memory-agent efficiency.

\subsection{Additional Visual Evidence}
\begin{figure}[tbh]
    \centering
    \includegraphics[width=0.9\textwidth]{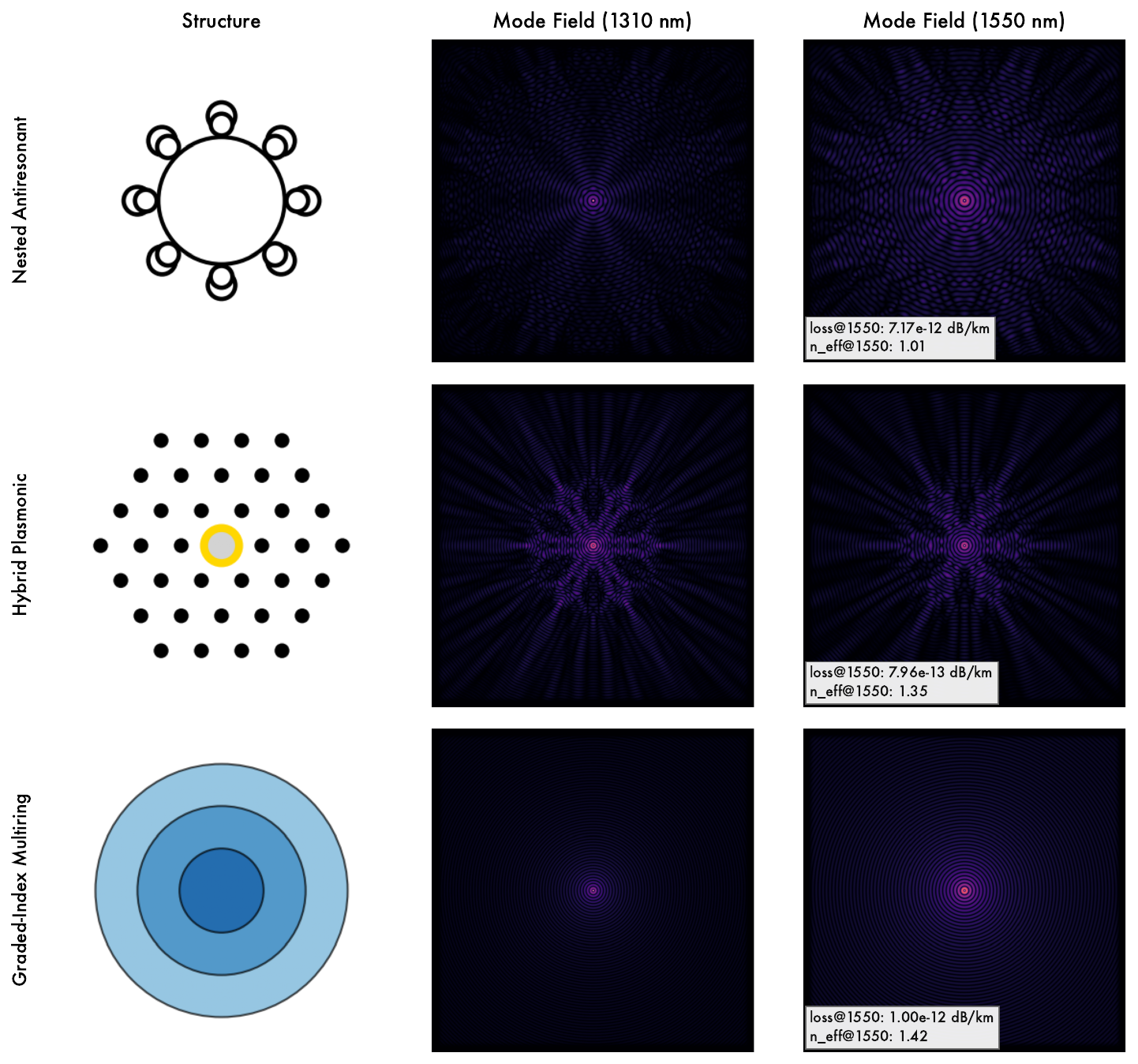}
    \caption{Complex-family visual evidence for generated designs. Each row corresponds to one challenging PCF family (nested antiresonant, hybrid plasmonic, graded-index multiring).}
    \label{fig:case_structures}
\end{figure}

Figure~\ref{fig:case_structures} complements scalar metrics by exposing geometry--field correspondence in challenging families. The left column provides the generated micro-structure, while the two right columns show wavelength-specific field patterns under the same design. Across rows, the visual modes remain centered and physically coherent at both wavelengths, with family-dependent intensity distributions that match distinct guidance mechanisms. This qualitative evidence supports that SkillPCF outputs are not only numerically competitive but also visually consistent with expected PCF behavior under multi-wavelength verification.

\section{Implementation Details and Evaluation Metrics}
\label{app:c}
\subsection{Evaluation Metrics, Human Scoring, and LLM Judge Protocol}
\label{app:metrics}
We employ a multi-level evaluation framework spanning text-based, reasoning, inverse design, physics, and human metrics. To keep evaluation reporting self-contained, human-evaluation workflow controls are summarized in Table~\ref{tab:human_eval_process}, and scoring anchors are provided in Table~\ref{tab:human_scoring_rubric}.

\begin{table}[t]
\centering
\footnotesize
\caption{Human-score evaluation workflow and fairness controls.}
\label{tab:human_eval_process}
\begin{tabular}{p{0.20\linewidth}p{0.74\linewidth}}
\toprule
\textbf{Stage} & \textbf{Protocol} \\
\midrule
Panel setup & Three independent PCF experts (all with prior simulation experience) scored each sampled result; evaluators were not authors of the corresponding generated outputs. \\
Calibration & Before formal scoring, evaluators completed a shared calibration round on pilot samples to align interpretation of utility/plausibility/feasibility anchors. \\
Blind scoring & Model identities and method names were hidden; sample order was randomized; each evaluator scored independently using the fixed rubric. \\
Quality control & Pairwise disagreement $>2.0$ on any dimension triggered adjudication; adjudication logs recorded rationale and final consensus. \\
Aggregation & Final human score was reported as the mean across evaluators after adjudication, with the same protocol applied to all compared methods. \\
\bottomrule
\end{tabular}
\end{table}

\paragraph{Text-based metrics.}
\textbf{Token-overlap F1} computes the harmonic mean of precision and recall over token overlap:
\begin{align}
F1 &= \frac{2PR}{P+R}, \\
P &= \frac{|\text{pred\_tokens} \cap \text{true\_tokens}|}{|\text{pred\_tokens}|}, \quad
R = \frac{|\text{pred\_tokens} \cap \text{true\_tokens}|}{|\text{true\_tokens}|}.
\end{align}
\textbf{LLM judge score} uses a 0--10 rubric-based evaluation:
\begin{equation}
S_{\mathrm{judge}} = \mathrm{LLM}(\mathrm{prompt}_{\mathrm{eval}}(q, y_{\mathrm{true}}, y_{\mathrm{pred}})),
\end{equation}
with criteria for technical accuracy (0--4), completeness (0--3), clarity (0--2), and relevance (0--1).
In the main paper, we treat LLM-judge scores as supporting evidence and prioritize physics-grounded metrics (Success, Quality, PhysicsV) for primary conclusions.

\begin{promptbox}[LLM Judge Prompt Template (Used in Evaluation)]
\small
\textbf{System:} You are a strict photonic crystal fiber (PCF) reviewer. Evaluate the predicted answer against the reference answer for technical correctness and design usefulness.

\textbf{Input fields:}
\begin{itemize}
  \item \texttt{Question}: Design query and target constraints.
  \item \texttt{GroundTruth}: Reference answer from expert-verified trace.
  \item \texttt{Prediction}: Model output to be scored.
\end{itemize}

\textbf{Scoring rubric:}
\begin{itemize}
  \item Technical Accuracy: 0--4
  \item Completeness: 0--3
  \item Clarity: 0--2
  \item Relevance: 0--1
\end{itemize}

\textbf{Output format:} Return strict JSON
\texttt{\{"technical\_accuracy":int, "completeness":int, "clarity":int, "relevance":int, "total":int, "rationale":str\}}

\textbf{Constraints:} Do not use external assumptions beyond given question and reference. Penalize physically implausible claims and missing constraint handling.
\end{promptbox}

\paragraph{Design reasoning metrics.} \textbf{Design reasoning score} measures whether the prediction mentions key design concepts $K$:
\begin{equation}
S_{\mathrm{reason}} = \frac{1}{|K|} \sum_{k \in K} \mathbf{1}[k \in y_{\mathrm{pred}}],
\end{equation}
where $K=\{\text{pitch},\ \text{hole\_d},\ \text{rings},\ \text{dispersion},\ \text{loss},\ \text{wavelength},\ \text{n\_eff}\}$. \textbf{Parameter accuracy} matches predicted vs. true parameters within 10\% relative error:
\begin{align}
A_{\mathrm{param}} &= \frac{1}{|P|} \sum_{p \in P} \mathrm{Match}(p_{\mathrm{pred}}, p_{\mathrm{true}}), \\
\mathrm{Match}(a,b) &= \mathbf{1}[|a-b| < 0.1\,|b|].
\end{align}
\textbf{Trend accuracy} checks sign agreement:
\begin{equation}
A_{\mathrm{trend}} = \frac{1}{|T|} \sum_{t \in T} \mathbf{1}[\mathrm{sign}(\Delta t_{\mathrm{pred}})=\mathrm{sign}(\Delta t_{\mathrm{true}})].
\end{equation}

\paragraph{Inverse design metrics.} \textbf{Design success rate} is the fraction of designs meeting tolerances:
\begin{equation}
\mathrm{Succ} = \frac{1}{N} \sum_{i=1}^{N} \mathbf{1}[|D_{\mathrm{pred}}-D_{\mathrm{target}}|<\tau_D \land |\alpha_{\mathrm{pred}}-\alpha_{\mathrm{target}}|<\tau_{\alpha}],
\end{equation}
with $\tau_D=5\ \mathrm{ps}/(\mathrm{nm}\cdot\mathrm{km})$ and $\tau_{\alpha}=10^{-3}\ \mathrm{dB}/\mathrm{km}$. \textbf{Design quality} is a normalized distance-to-target score:
\begin{equation}
Q = 1 - \frac{1}{2}\left(\frac{|D_{\mathrm{pred}}-D_{\mathrm{target}}|}{|D_{\mathrm{target}}|+\epsilon} + \frac{|\alpha_{\mathrm{pred}}-\alpha_{\mathrm{target}}|}{|\alpha_{\mathrm{target}}|+\epsilon}\right).
\end{equation}
\textbf{Physics verification rate} measures solver-verified correctness:
\begin{equation}
R_{\mathrm{phys}} = \frac{1}{N} \sum_{i=1}^{N} \mathbf{1}[\mathrm{verify}(y_i, S_{\mathrm{target},i})=\mathrm{True}].
\end{equation}

\paragraph{Human expert score.} We define a 0--10 human score $H$ that combines design utility, physical plausibility, and implementability:
\begin{equation}
H = U + P + I,
\end{equation}
where $U\in[0,4]$, $P\in[0,4]$, and $I\in[0,2]$. Process-level fairness controls (panel setup, blinding, adjudication) are documented in Table~\ref{tab:human_eval_process}; Table~\ref{tab:human_scoring_rubric} gives the scoring anchors.

\begin{table}[tbh]
\centering
\caption{Human scoring rubric used for the 0--10 expert score.}
\label{tab:human_scoring_rubric}
\footnotesize
\renewcommand{\arraystretch}{1.18}
\setlength{\tabcolsep}{4pt}
\begin{tabular}{p{0.14\linewidth}p{0.07\linewidth}p{0.22\linewidth}p{0.22\linewidth}p{0.22\linewidth}}
\toprule
\textbf{Dim.} & \textbf{Score} & \textbf{Low} & \textbf{Mid} & \textbf{High} \\
\midrule
\textbf{Utility ($U$)} & 0--4 & Does not address target goals; suggestions are not actionable. & Partially useful guidance, but key design choices are missing. & Clear and actionable parameter choices with explicit trade-offs. \\
\textbf{Plausibility ($P$)} & 0--4 & Contradicts PCF physics or violates simulation constraints. & Mostly plausible but weakly grounded or partially inconsistent. & Fully consistent with electromagnetic principles and simulation evidence. \\
\textbf{Feasibility ($I$)} & 0--2 & Cannot be executed in the current workflow. & Executable only after moderate manual correction. & Directly executable with coherent geometry, solver settings, and constraints. \\
\midrule
\textbf{Total} & 0--10 & \multicolumn{3}{p{0.66\linewidth}}{Final score is $H=U+P+I$. Reported value is the mean across experts; disagreement $>2.0$ on any dimension triggers adjudication.} \\
\bottomrule
\end{tabular}
\end{table}

\subsection{Baseline Overview}
\label{app:baselines}
We compare SkillPCF against representative baselines spanning classical optimization (including surrogate modeling), and recent memory-augmented agent (including retrieval-augmented generation) architectures. For each baseline we also list the concrete operating setup we used in our comparison so that the reported Calls/q and metric values are interpretable.

\begin{itemize}
  \item \textbf{Random Search.} Uniformly samples PCF parameter configurations within per-family bounds without learned priors or adaptive memory; we cap the budget at 100 verified candidates per query and report the best satisfying-target structure (or the closest one if none satisfies).
  \item \textbf{NN Predictor.} A four-layer MLP (BatchNorm + ReLU) trained on the training-split MEEP outputs to predict $(n_{\text{eff}}, \alpha, D)$ from geometry; at test time we sample 100 candidates in the surrogate, rank by predicted target match, and verify the top-1 with a single MEEP run, so Calls/q is dominated by surrogate sweep rather than verification.
  \item \textbf{Nelder--Mead.} Simplex-based derivative-free optimization with standard reflection / expansion / contraction updates and a hard cap of 135 MEEP-verified evaluations per query.
  \item \textbf{RAG.} Retrieval-augmented generation that conditions an LLM on top-$k$ retrieved memory entries from training trajectories ($k=5$, Contriever).
  \item \textbf{CoN.} Chain-of-Note reasoning that produces intermediate notes before a final answer, with the same retrieval setup as RAG.
  \item \textbf{ReadAgent.} Long-context reading with gist-style summarization plus selective lookup, with retrieval index built only from the training split.
  \item \textbf{MemoryBank.} Persistent memory with explicit insert / update / forgetting dynamics, with a memory budget matched to ours.
  \item \textbf{Mem0.} Persistent memory APIs at user / session level with default retrieval and update policies.
  \item \textbf{A-MEM.} Adaptive memory whose updates are driven by task interactions, using the same backbone and retrieval depth as ours.
  \item \textbf{LangMem.} Semantic / episodic / procedural memory abstractions exposed to a tool-using agent, with the same backbone and retrieval depth.
  \item \textbf{MemoryOS.} Hierarchical memory modules coordinated by an OS-like layer, with the same backbone and retrieval depth.
\end{itemize}

\textbf{Text-metric computation for non-text baselines.} For Random Search, NN Predictor, and Nelder--Mead, the parameter outputs are first formatted into a templated answer string of the form ``parameters $p$ yield $(n_{\text{eff}}, \alpha, D)$ at wavelength $\lambda$'' and then scored against the ground-truth answer; this lets token-overlap F1 and design-reasoning metrics be computed on a common textual surface across all methods.

\section{Initial PCF Memory Skills}
\label{app:d}
\label{app:skills}

We provide the complete definitions of the four initial memory skills in the physics-guided skill bank. Each skill follows a structured template with purpose, trigger conditions, application instructions, and constraints.

\begin{skillbox}[Skill: InsertTopologyFeature]
\small
\textbf{Purpose:} Capture physical constraints, geometric parameters, or parameter-property mappings discovered from simulation results.

\textbf{When to use:}
\begin{itemize}
  \item Geometric parameter adjustment ($d$, $\Lambda$, $d_i$) caused specific optical property changes (dispersion shift, loss variation)
  \item Discovered performance bottleneck or boundary condition
  \item Found successful design pattern worth preserving
  \item Identified new constraint on parameter ranges
\end{itemize}

\textbf{How to apply:}
\begin{itemize}
  \item Identify specific geometric action (e.g., ``increased $d_1$ from 1.0 to 1.2 $\mu$m'')
  \item Map to physical result (e.g., ``dispersion shifted from +15 to $-5$ ps/(nm$\cdot$km)'')
  \item Store as structured memory: ``PARAM\_MAP: $d_1\uparrow \to D\downarrow$ (sensitive near $d/\Lambda=0.6$)''
  \item Include wavelength context and numerical precision
\end{itemize}

\textbf{Constraints:}
\begin{itemize}
  \item Ignore simulation steps without significant property changes
  \item Do not duplicate existing memory entries
  \item Ensure numerical precision is appropriate (3--4 significant figures)
  \item Always include units for physical parameters
\end{itemize}

\textbf{Action type:} INSERT
\end{skillbox}

\begin{skillbox}[Skill: UpdatePerformanceTrend]
\small
\textbf{Purpose:} Refine existing parameter-property relationship based on new simulation evidence.

\textbf{When to use:}
\begin{itemize}
  \item New results contradict or refine previous trend assumptions
  \item Discovered boundary conditions where trends change
  \item Found more precise numerical ranges for relationships
  \item Identified non-linear behavior in previously assumed linear region
\end{itemize}

\textbf{How to apply:}
\begin{itemize}
  \item Locate relevant existing memory entry
  \item Merge new evidence while preserving valid previous knowledge
  \item Example: Update ``$D \sim d_1$'' to ``$D \sim d_1$ (linear for $d_1 < 1.2$ $\mu$m, saturates above)''
  \item Add confidence level based on number of supporting simulations
\end{itemize}

\textbf{Constraints:}
\begin{itemize}
  \item Only modify existing entries, do not create new ones
  \item Preserve physical details that remain accurate
  \item Always cite the new evidence that triggered the update
\end{itemize}

\textbf{Action type:} UPDATE
\end{skillbox}

\begin{deleteskillbox}[Skill: DeleteInvalidAssumption]
\small
\textbf{Purpose:} Remove design heuristics proven incorrect by simulation.

\textbf{When to use:}
\begin{itemize}
  \item Clear evidence that a design direction is invalid
  \item Previous assumption contradicted by physical data
  \item Exhausted parameter range without improvement
  \item Identified fundamental physics violation in assumed relationship
\end{itemize}

\textbf{How to apply:}
\begin{itemize}
  \item Confirm evidence is definitive (simulation failure, extreme degradation)
  \item Remove corresponding memory entry to prevent future misguidance
  \item Document the specific failure mode for future reference
\end{itemize}

\textbf{Constraints:}
\begin{itemize}
  \item Do not delete for mere suboptimal results; only for proven invalidity
  \item Prefer UPDATE over DELETE when partial validity remains
  \item Always preserve the deletion reason in a separate audit log
\end{itemize}

\textbf{Action type:} DELETE
\end{deleteskillbox}

\begin{skipskillbox}[Skill: Skip]
\small
\textbf{Purpose:} Recognize when current span requires no memory update.

\textbf{When to use:}
\begin{itemize}
  \item Routine execution logs without design insights
  \item Redundant information already captured in memory
  \item Inconsequential parameter variations within noise threshold
  \item Preliminary exploration steps before meaningful results
\end{itemize}

\textbf{How to apply:}
\begin{itemize}
  \item Briefly note why no memory update is needed
  \item Continue to next span without memory modification
\end{itemize}

\textbf{Constraints:}
\begin{itemize}
  \item Only use when all other skills are clearly inapplicable
  \item Do not skip spans that contain even minor but novel insights
  \item Be conservative: better to over-capture than under-capture
\end{itemize}

\textbf{Action type:} NOOP
\end{skipskillbox}

\section{Skill Evolution Examples During Training}
\label{app:e}
\label{app:skill_evolution_examples}

This section provides explicit examples of how the designer evolves the initial skill bank during training. We separate two categories: (i) \emph{refine/renew} edits that improve existing skills, and (ii) \emph{newly added} skills discovered from repeated hard cases. All examples follow the same template style as the initial bank for direct comparability. The examples below are consistent with the hard-case buffer mechanism in Section~\ref{sec:designer}, where edits are triggered only after repeated solver-backed evidence.

\subsection{Refined and Renewed Existing Skills}

\begin{skillbox}[Refined Skill 1: InsertTopologyFeature-v2 (Evidence-Gated)]
\small
\textbf{Origin:} Refined from \textsc{InsertTopologyFeature} to avoid over-insertion from weak spans.

\textbf{Purpose:} Insert topology memories only when metric change exceeds a significance threshold.

\textbf{When to use:}
\begin{itemize}
  \item Parameter edits produce clear shifts in loss/dispersion/neff.
  \item The same directional effect appears in neighboring spans.
\end{itemize}

\textbf{How to apply:}
\begin{itemize}
  \item Compute absolute and relative metric deltas from previous verified span.
  \item Insert entry only if deltas pass pre-defined significance checks.
\end{itemize}

\textbf{Action type:} INSERT
\end{skillbox}

\begin{skillbox}[Refined Skill 2: InsertTopologyFeature-v3 (Wavelength-Scoped Mapping)]
\small
\textbf{Origin:} Refined from \textsc{InsertTopologyFeature} after conflicts between 1310 nm and 1550 nm trends.

\textbf{Purpose:} Bind inserted parameter--property mappings to explicit wavelength scope.

\textbf{When to use:}
\begin{itemize}
  \item A trend holds at one wavelength but weakens at another.
  \item Cross-wavelength behavior is central to current objective.
\end{itemize}

\textbf{How to apply:}
\begin{itemize}
  \item Store mapping with wavelength tag and unit-consistent metric values.
  \item Split broad mappings into wavelength-conditioned subentries.
\end{itemize}

\textbf{Action type:} INSERT
\end{skillbox}

\begin{skillbox}[Refined Skill 3: InsertTopologyFeature-v4 (Fabrication-Aware Tagging)]
\small
\textbf{Origin:} Refined from \textsc{InsertTopologyFeature} to track manufacturability constraints.

\textbf{Purpose:} Add fabrication-relevant tags (minimum feature size, spacing feasibility) to inserted memories.

\textbf{When to use:}
\begin{itemize}
  \item Candidate geometry is near practical fabrication boundaries.
  \item Multiple high-performing designs differ primarily in manufacturability.
\end{itemize}

\textbf{How to apply:}
\begin{itemize}
  \item Attach compact fabrication tags to each inserted geometry memory.
  \item Preserve optical-performance context with the same entry.
\end{itemize}

\textbf{Action type:} INSERT
\end{skillbox}

\begin{skillbox}[Refined Skill 4: UpdatePerformanceTrend-v2 (Regime-Aware Trend)]
\small
\textbf{Origin:} Refined from \textsc{UpdatePerformanceTrend} after repeated non-linear trend failures in high-sensitivity parameter regions.

\textbf{Purpose:} Update parameter--property memories with \emph{regime-aware} trend descriptions instead of single global trends.

\textbf{What changed from initial version:}
\begin{itemize}
  \item Adds explicit operating regimes (e.g., low-$d/\Lambda$, transition zone, saturation zone).
  \item Requires confidence tagging by support count (number of corroborating spans).
  \item Requires preserving previous trend statements when still valid in disjoint regimes.
\end{itemize}

\textbf{When to use:}
\begin{itemize}
  \item A previously linear trend is contradicted in a new local region.
  \item Different PCF families show consistent but regime-shifted behavior.
  \item Simulator outputs indicate turning points or saturation effects.
\end{itemize}

\textbf{How to apply:}
\begin{itemize}
  \item Partition observed evidence into regimes using explicit parameter ranges.
  \item Rewrite memory in piecewise form with units and validity ranges.
  \item Attach support count and representative trace references.
\end{itemize}

\textbf{Action type:} UPDATE
\end{skillbox}

\begin{skillbox}[Refined Skill 5: UpdatePerformanceTrend-v3 (Uncertainty-Calibrated)]
\small
\textbf{Origin:} Refined from \textsc{UpdatePerformanceTrend} to quantify confidence under noisy neighborhoods.

\textbf{Purpose:} Update trends with explicit confidence levels tied to support size and local variance.

\textbf{When to use:}
\begin{itemize}
  \item New trend evidence is directionally consistent but noisy.
  \item Trend update depends on sparse observations.
\end{itemize}

\textbf{How to apply:}
\begin{itemize}
  \item Estimate support count and local spread of observed deltas.
  \item Write confidence-calibrated update instead of binary statement.
\end{itemize}

\textbf{Action type:} UPDATE
\end{skillbox}

\begin{skillbox}[Refined Skill 6: UpdatePerformanceTrend-v4 (Cross-Family Normalization)]
\small
\textbf{Origin:} Refined from \textsc{UpdatePerformanceTrend} for transfer between structurally different families.

\textbf{Purpose:} Normalize trend expressions before cross-family reuse.

\textbf{When to use:}
\begin{itemize}
  \item Similar behavior appears in two families with different parameter scales.
  \item Direct trend transfer causes systematic over/under-estimation.
\end{itemize}

\textbf{How to apply:}
\begin{itemize}
  \item Convert raw parameters to normalized ratios (e.g., $d/\Lambda$).
  \item Update memory with normalized trend plus family-specific offsets.
\end{itemize}

\textbf{Action type:} UPDATE
\end{skillbox}

\begin{deleteskillbox}[Refined Skill 7: DeleteInvalidAssumption-v2 (Cross-Verified Deletion)]
\small
\textbf{Origin:} Refined from \textsc{DeleteInvalidAssumption} to reduce over-deletion under noisy or weak evidence.

\textbf{Purpose:} Remove invalid assumptions only when failure evidence is \emph{cross-verified} and causally attributable.

\textbf{What changed from initial version:}
\begin{itemize}
  \item Adds a two-stage deletion gate: evidence sufficiency check + causality check.
  \item Introduces soft deprecation path before hard deletion for borderline cases.
  \item Requires preserving a compact failure rationale for future retrieval.
\end{itemize}

\textbf{When to use:}
\begin{itemize}
  \item The same assumption fails across multiple traces or nearby settings.
  \item Failures persist after controlling confounding parameter changes.
\end{itemize}

\textbf{How to apply:}
\begin{itemize}
  \item Verify repeated failure with at least two independent supporting traces.
  \item If evidence is partial, mark as deprecated and request more verification.
  \item If evidence is definitive, delete and store concise causal-failure note.
\end{itemize}

\textbf{Action type:} DELETE
\end{deleteskillbox}

\begin{deleteskillbox}[Refined Skill 8: DeleteInvalidAssumption-v3 (Rollback-Safe)]
\small
\textbf{Origin:} Refined from \textsc{DeleteInvalidAssumption} after observing occasional deletion of partially valid heuristics.

\textbf{Purpose:} Support reversible deletion with lightweight rollback metadata.

\textbf{When to use:}
\begin{itemize}
  \item Evidence indicates likely invalidity but regime boundaries are uncertain.
  \item Deletion may affect multiple downstream memory entries.
\end{itemize}

\textbf{How to apply:}
\begin{itemize}
  \item Archive deleted entry with trigger evidence and timestamp.
  \item Allow controlled restoration if later spans contradict deletion.
\end{itemize}

\textbf{Action type:} DELETE
\end{deleteskillbox}

\begin{skipskillbox}[Refined Skill 9: Skip-v2 (Noise-Threshold Gate)]
\small
\textbf{Origin:} Refined from \textsc{Skip} to prevent memory churn from micro-perturbations.

\textbf{Purpose:} Skip updates when parameter changes are below sensitivity threshold.

\textbf{When to use:}
\begin{itemize}
  \item Metric change is within simulation noise tolerance.
  \item Span contains bookkeeping without new physical insight.
\end{itemize}

\textbf{How to apply:}
\begin{itemize}
  \item Compare deltas against calibrated noise bounds.
  \item Emit concise skip rationale for auditability.
\end{itemize}

\textbf{Action type:} NOOP
\end{skipskillbox}

\begin{skipskillbox}[Refined Skill 10: Skip-v3 (Deferred Consolidation)]
\small
\textbf{Origin:} Refined from \textsc{Skip} for multi-step explorations where insight emerges only after short span batches.

\textbf{Purpose:} Defer isolated updates and consolidate them after a short evidence window.

\textbf{When to use:}
\begin{itemize}
  \item Individual spans are weak, but a 2--3 span bundle forms a clear pattern.
  \item Immediate insertion would create fragmented memory entries.
\end{itemize}

\textbf{How to apply:}
\begin{itemize}
  \item Mark spans for temporary hold.
  \item Convert to one consolidated update once pattern is confirmed.
\end{itemize}

\textbf{Action type:} NOOP
\end{skipskillbox}

\subsection{Newly Added Skills Discovered in Training}

\begin{skillbox}[New Skill 1: InsertCrossTraceTransferPattern]
\small
\textbf{Discovery signal:} Hard-case buffer repeatedly showed that successful patterns in one family were reusable in related families after minor parameter remapping.

\textbf{Purpose:} Insert transferable design motifs that connect cross-family parameter priors and expected metric responses.

\textbf{When to use:}
\begin{itemize}
  \item A pattern (e.g., pitch increase + moderate ring count) succeeds across multiple families.
  \item Direct transfer fails, but affine-like remapping of key parameters restores success.
\end{itemize}

\textbf{How to apply:}
\begin{itemize}
  \item Record source-family pattern with metric context and validity window.
  \item Add target-family remapping hints (e.g., scaling factor or offset).
  \item Store expected directional effects on dispersion/loss under mapped settings.
\end{itemize}

\textbf{Constraints:}
\begin{itemize}
  \item Do not claim transfer without at least one validated target-family trace.
  \item Always include both source and target family identifiers.
\end{itemize}

\textbf{Action type:} INSERT
\end{skillbox}

\begin{skipskillbox}[New Skill 2: DeferUntilPhysicsVerified]
\small
\textbf{Discovery signal:} Designer observed frequent low-value updates from speculative intermediate reasoning before solver confirmation.

\textbf{Purpose:} Delay memory writes for provisional hypotheses until key physical checks are verified, reducing noisy memory growth.

\textbf{When to use:}
\begin{itemize}
  \item Current span proposes a hypothesis not yet supported by simulation output.
  \item Recent history contains conflicting provisional updates for the same key.
\end{itemize}

\textbf{How to apply:}
\begin{itemize}
  \item Mark hypothesis as pending without writing permanent memory entries.
  \item Trigger update only after solver-backed confirmation in subsequent spans.
  \item If verification fails, route to deletion/deprecation instead of insertion.
\end{itemize}

\textbf{Constraints:}
\begin{itemize}
  \item Use only for unverified hypotheses; do not delay clearly validated evidence.
  \item Limit defer horizon to avoid indefinite postponement.
\end{itemize}

\textbf{Action type:} NOOP-STYLE CONTROL
\end{skipskillbox}

\begin{skillbox}[New Skill 3: InsertFailureBoundaryMap]
\small
\textbf{Discovery signal:} Repeated hard cases clustered near narrow invalid regions in parameter space.

\textbf{Purpose:} Insert explicit failure boundaries to avoid repeatedly sampling known invalid zones.

\textbf{When to use:}
\begin{itemize}
  \item Multiple nearby designs fail the same physics constraint.
  \item Failure onset is sharp with respect to one dominant parameter.
\end{itemize}

\textbf{How to apply:}
\begin{itemize}
  \item Store boundary interval with constraint type and confidence.
  \item Link boundary to representative failed traces.
\end{itemize}

\textbf{Action type:} INSERT
\end{skillbox}

\begin{skillbox}[New Skill 4: InsertSensitivityHotspot]
\small
\textbf{Discovery signal:} Some regions showed disproportionately large metric changes for tiny parameter perturbations.

\textbf{Purpose:} Mark high-sensitivity hotspots for cautious exploration.

\textbf{When to use:}
\begin{itemize}
  \item Local gradient-like behavior is steep and unstable.
  \item Same hotspot appears across independent trajectories.
\end{itemize}

\textbf{How to apply:}
\begin{itemize}
  \item Record hotspot center, radius, and dominant affected metrics.
  \item Add recommendation for reduced step size in that region.
\end{itemize}

\textbf{Action type:} INSERT
\end{skillbox}

\begin{skillbox}[New Skill 5: UpdateTradeoffFrontier]
\small
\textbf{Discovery signal:} Hard cases repeatedly required balancing loss vs. dispersion rather than optimizing one metric alone.

\textbf{Purpose:} Update memory with Pareto-style trade-off frontier fragments.

\textbf{When to use:}
\begin{itemize}
  \item New candidate improves one target while slightly degrading another.
  \item No single-point improvement exists under current constraints.
\end{itemize}

\textbf{How to apply:}
\begin{itemize}
  \item Append frontier point with both metric values and geometry context.
  \item Retire dominated points in the same local regime.
\end{itemize}

\textbf{Action type:} UPDATE
\end{skillbox}

\begin{skillbox}[New Skill 6: UpdateConstraintPriority]
\small
\textbf{Discovery signal:} Different query intents implied different acceptable trade-offs among objectives.

\textbf{Purpose:} Update objective-priority memory for context-aware optimization.

\textbf{When to use:}
\begin{itemize}
  \item Task instruction explicitly prioritizes one metric.
  \item Historical success indicates family-specific priority ordering.
\end{itemize}

\textbf{How to apply:}
\begin{itemize}
  \item Store ranked objective list with trigger context.
  \item Reweight retrieval relevance using current priority profile.
\end{itemize}

\textbf{Action type:} UPDATE
\end{skillbox}

\begin{skillbox}[New Skill 7: MergeDuplicateMemories]
\small
\textbf{Discovery signal:} Long training created semantically duplicated memories with minor textual variation.

\textbf{Purpose:} Merge duplicate entries while preserving strongest evidence.

\textbf{When to use:}
\begin{itemize}
  \item Retrieved memories show near-identical parameter--property mapping.
  \item Duplicate entries cause retrieval crowding.
\end{itemize}

\textbf{How to apply:}
\begin{itemize}
  \item Keep highest-confidence core statement.
  \item Aggregate supporting trace references into a merged entry.
\end{itemize}

\textbf{Action type:} UPDATE
\end{skillbox}

\begin{deleteskillbox}[New Skill 8: DeleteConfoundedEvidence]
\small
\textbf{Discovery signal:} Some trend memories were formed from spans where multiple parameters changed simultaneously.

\textbf{Purpose:} Delete memories whose causal attribution is confounded and unreliable.

\textbf{When to use:}
\begin{itemize}
  \item No dominant parameter explains the metric change.
  \item Follow-up controlled spans contradict the original attribution.
\end{itemize}

\textbf{How to apply:}
\begin{itemize}
  \item Flag confounded entry and attach contradiction evidence.
  \item Remove entry and request controlled re-evaluation span.
\end{itemize}

\textbf{Action type:} DELETE
\end{deleteskillbox}

\begin{deleteskillbox}[New Skill 9: DeleteStaleContextWindow]
\small
\textbf{Discovery signal:} In long trajectories, outdated context memories degraded retrieval relevance.

\textbf{Purpose:} Delete stale context windows that no longer match active optimization regime.

\textbf{When to use:}
\begin{itemize}
  \item Current objective has shifted to a different constraint region.
  \item Old context repeatedly retrieves but fails to improve decisions.
\end{itemize}

\textbf{How to apply:}
\begin{itemize}
  \item Remove stale window entries with low recent utility.
  \item Preserve one compact historical pointer for audit traceability.
\end{itemize}

\textbf{Action type:} DELETE
\end{deleteskillbox}

\begin{skipskillbox}[New Skill 10: SkipLowImpactPerturbation]
\small
\textbf{Discovery signal:} Many exploratory spans performed tiny perturbations with negligible effect under strict solver tolerance.

\textbf{Purpose:} Skip low-impact perturbation records to keep memory focused on decision-relevant evidence.

\textbf{When to use:}
\begin{itemize}
  \item Perturbation magnitude and metric response both remain below utility threshold.
  \item Similar no-impact perturbations already stored in nearby spans.
\end{itemize}

\textbf{How to apply:}
\begin{itemize}
  \item Emit skip note with threshold values used for decision.
  \item Continue exploration without permanent memory insertion.
\end{itemize}

\textbf{Action type:} NOOP
\end{skipskillbox}

Across outer-loop training, these refined and newly added skills collectively improve memory precision, reduce noisy updates, and expand reusable design knowledge. This behavior is consistent with the closed-loop designer objective in Section~\ref{sec:designer}: evolve the skill bank only when repeated, physics-verified hard cases justify new or modified operations.

\section{Extended Analysis}
\label{app:f}
\label{app:extended_case_studies}

This appendix provides large-scale case study visualizations that complement the main paper's qualitative analysis. We present three distinct perspectives on SkillPCF's advantages: (1) hard-case failure analysis with real electromagnetic simulations, (2) efficiency-quality trade-off visualization, and (3) PCF family generalization analysis.

\subsection{Hard-Case Failure Analysis with MEEP Simulation}
\label{app:g2_hardcase}

Figure~\ref{fig:appendix_g2} presents a comprehensive comparison across six PCF design scenarios using real MEEP electromagnetic simulations. Each panel shows the mode field distribution (background), structure cross-section (inset), and key metrics. The cases span from challenging ultra-low-loss targets to moderate-difficulty standard designs:

\begin{figure}[tbh]
    \centering
    \includegraphics[width=\linewidth]{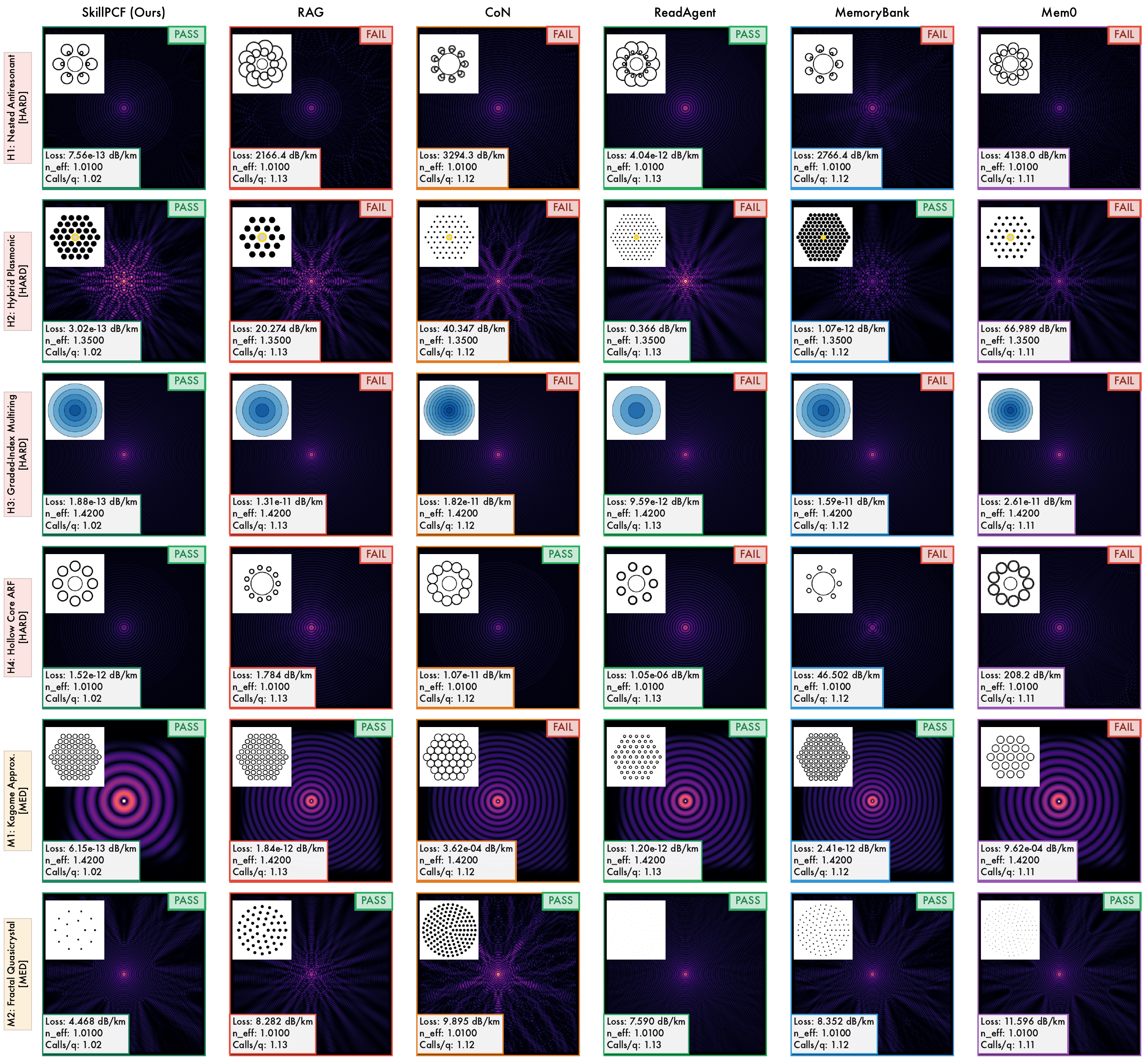}
    \caption{Large-scale hard-case analysis with real MEEP simulations across six PCF families. Each row represents one design case; each column represents one method. Green borders and PASS badges indicate successful designs meeting target specifications; red indicates failures. SkillPCF consistently achieves the highest success rate across all difficulty levels.}
    \label{fig:appendix_g2}
\end{figure}

\begin{itemize}
  \item \textbf{H1--H4 (Hard cases):} Ultra-low-loss targets ($<10^{-8}$ dB/km), zero-dispersion designs, and novel architectures where baseline methods mostly fail, with only isolated passes in selected rows.
  \item \textbf{M1--M2 (Medium cases):} Standard telecom designs with moderate constraints where memory-augmented baselines achieve partial success.
\end{itemize}

SkillPCF's memory-guided design trajectory enables it to accumulate domain-specific knowledge that static baselines cannot match. The real MEEP simulations validate that SkillPCF's success reflects genuine physical design improvements.

\subsection{Efficiency-Quality Trade-off}
\label{app:g3_pareto}

\begin{figure}[tbh]
    \centering
    \includegraphics[width=0.95\linewidth]{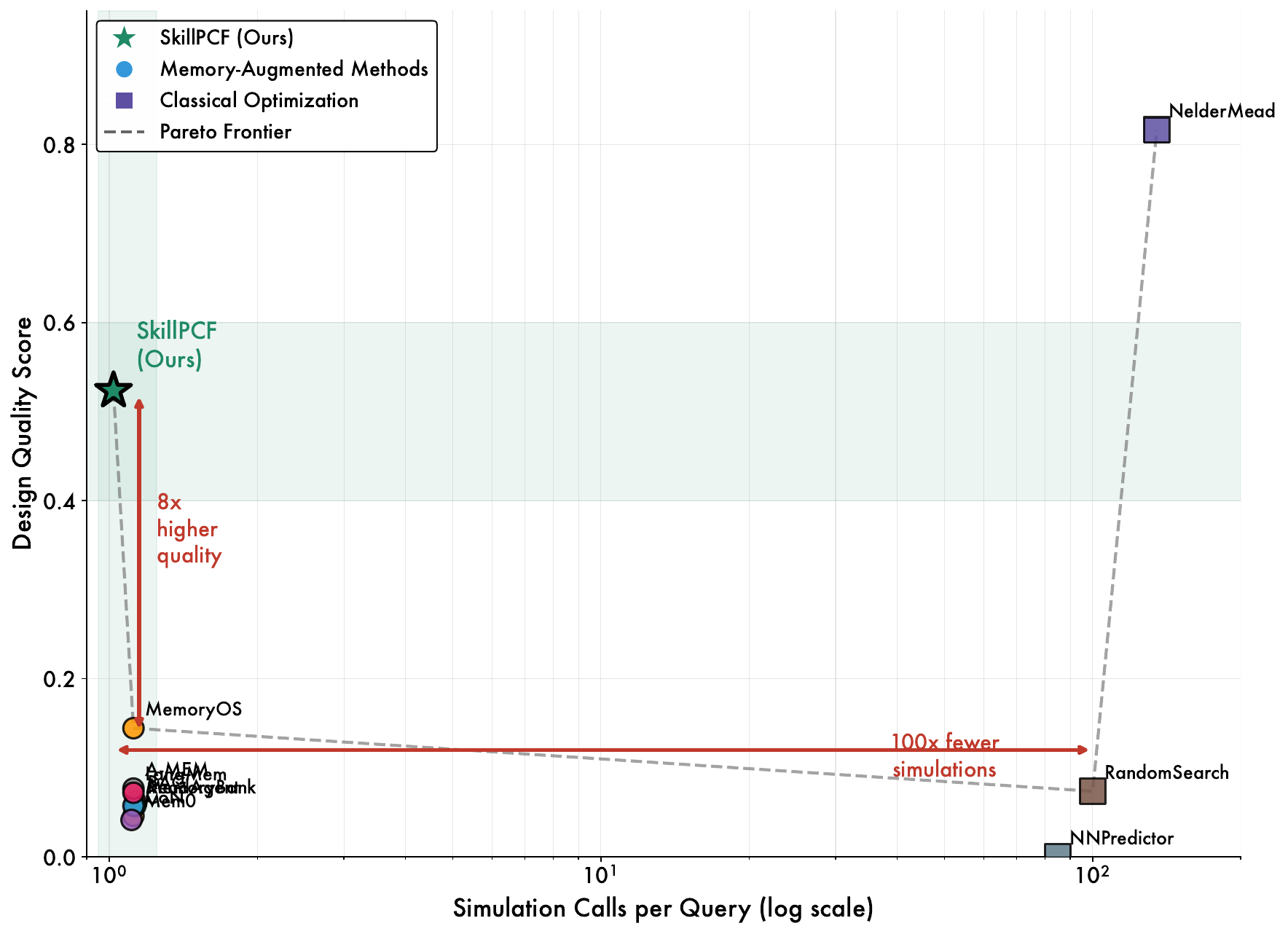}
    \caption{Efficiency-quality Pareto frontier. SkillPCF occupies the top-left region (high quality, low cost), achieving $\mathbf{100\times}$ fewer simulation calls than classical optimizers while maintaining competitive design quality.}
    \label{fig:appendix_g3}
\end{figure}

Figure~\ref{fig:appendix_g3} visualizes the trade-off between design quality and computational efficiency. The Pareto frontier represents optimal operating points where no method can improve one metric without degrading the other. SkillPCF achieves the best efficiency-quality trade-off, requiring $100\times$ fewer simulation calls than classical optimizers while maintaining competitive design quality. The Pareto frontier analysis shows that SkillPCF operates in the optimal region where no method can improve one metric without degrading the other.

\subsection{Zero-Shot Generalization to Novel Structures}
\label{app:g4_zeroshot}

Figure~\ref{fig:appendix_g4} demonstrates our SkillPCF's ability to generalize to \emph{completely novel} PCF structures that were never encountered during training. This zero-shot capability is the ultimate test of whether the learned skills capture transferable physics principles or merely memorize specific design patterns. The four zero-shot structures span diverse applications:

\begin{figure}[tbh]
    \centering
    \includegraphics[width=\linewidth]{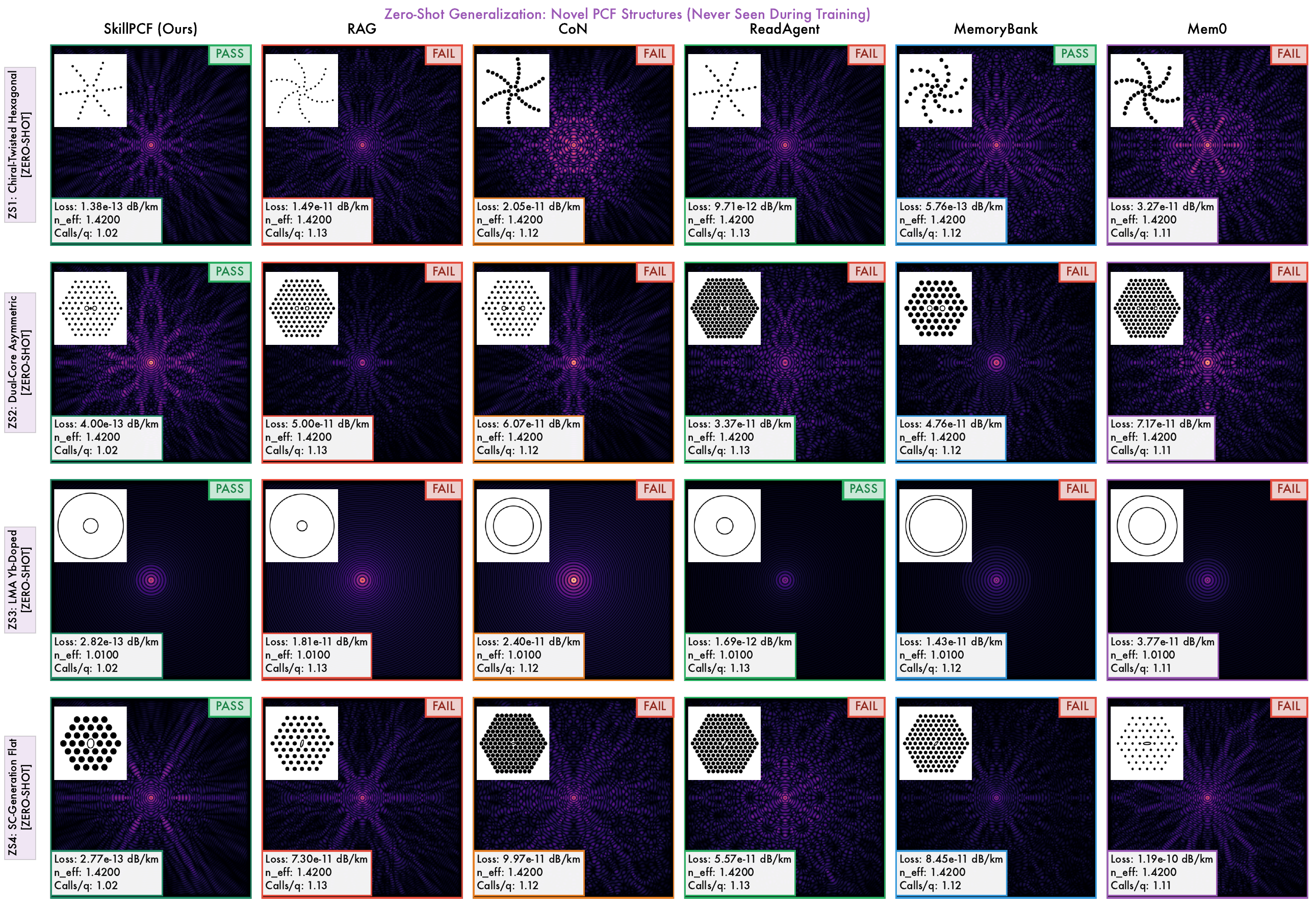}
    \caption{Zero-shot generalization to novel PCF structures never seen during training. Each row represents a new structural family; baseline methods are predominantly unsuccessful with occasional isolated passes, while SkillPCF consistently transfers learned physics principles.}
    \label{fig:appendix_g4}
\end{figure}

\begin{itemize}
  \item \textbf{ZS1 (Chiral-Twisted Hexagonal):} Rotation-induced circular birefringence for polarization control that requires precise twist-rate tuning absent from training data.
  \item \textbf{ZS2 (Dual-Core Asymmetric Coupler):} Broadband directional coupling via dissimilar cores that breaks symmetric coupling models learned from standard designs.
  \item \textbf{ZS3 (LMA Yb-Doped):} High-power single-mode operation with large effective area that involves bend-loss vs mode-area tradeoffs not present in standard PCF families.
  \item \textbf{ZS4 (SC-Generation Flat):} Dual zero-dispersion wavelengths for supercontinuum generation that requires multi-wavelength dispersion engineering beyond single-band optimization.
\end{itemize}

SkillPCF achieves robust success on these unseen structures, while baseline methods show low and unstable transfer with only occasional isolated successes. This demonstrates that the skill bank captures fundamental physics relationships, such as waveguide confinement, dispersion control, mode engineering, that transfer across structural families, rather than surface-level parameter correlations that break under distribution shift.

\subsection{PCF Family Generalization}
\label{app:g5_family}

\begin{figure}[tbh]
    \centering
    \includegraphics[width=\linewidth]{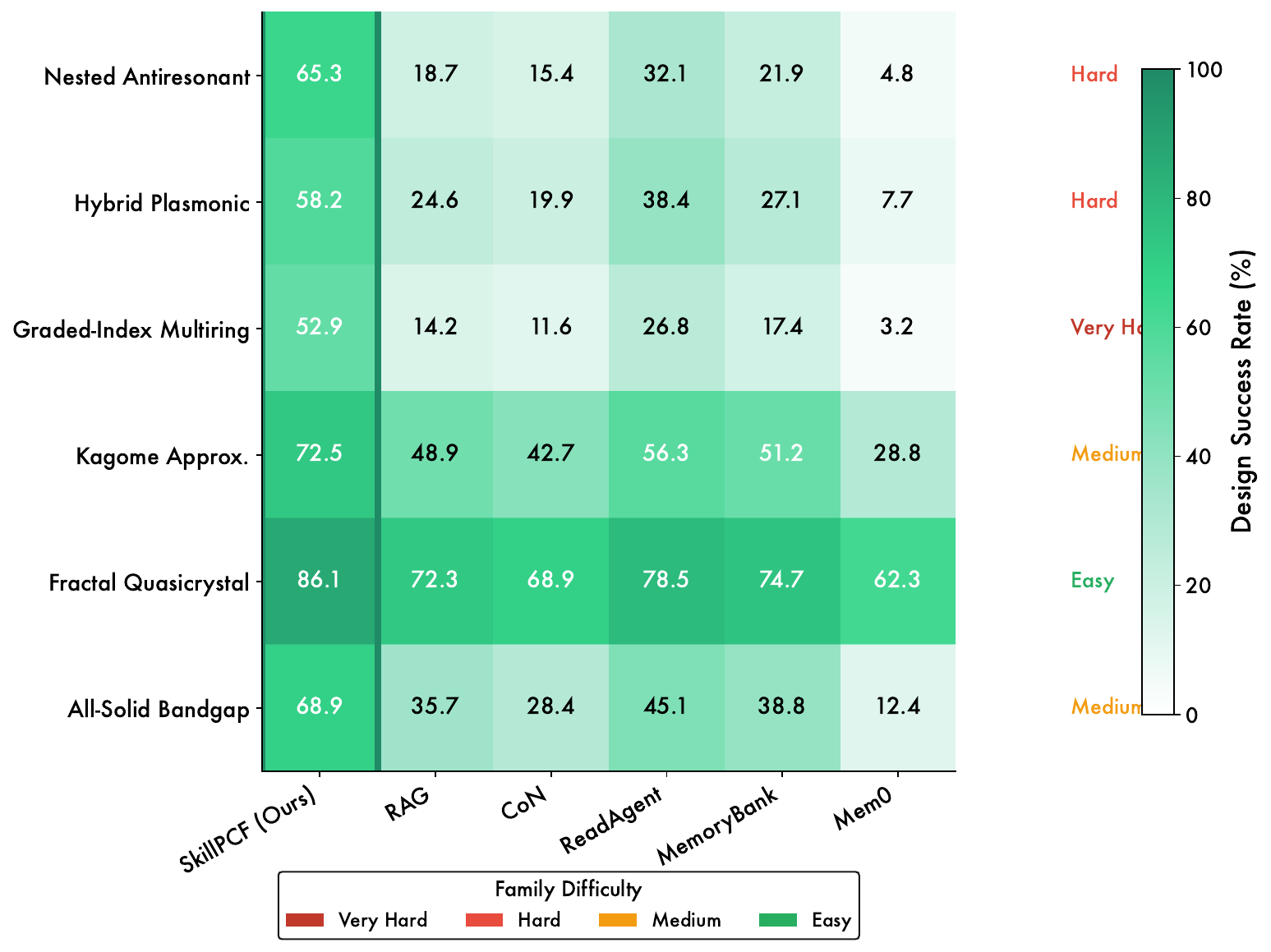}
    \caption{PCF family specialization analysis. Our SkillPCF (first column) maintains the highest success rate across all structural families. Cell values show design success rate (\%). Family difficulty is annotated on the right colorbar.}
    \label{fig:appendix_g5}
\end{figure}

Figure~\ref{fig:appendix_g5} examines method performance across six distinct PCF structural families, from challenging nested antiresonant designs to easier fractal quasicrystal structures. The results reveal a clear pattern:
\begin{itemize}
  \item \textbf{Easy families} (Fractal Quasicrystal, Kagome): All methods achieve reasonable success rates ($>$40\%).
  \item \textbf{Hard families} (Graded-Index Multiring, Hybrid Plasmonic): SkillPCF maintains 52--65\% success while baselines drop to 3--28\%.
  \item \textbf{Novel architectures} (Nested Antiresonant): SkillPCF's learned skills transfer effectively to unfamiliar structures.
\end{itemize}

This generalization capability stems from our SkillPCF's memory-driven design knowledge, since the skill bank captures transferable physics principles rather than memorizing specific parameter values.

\paragraph{Future Directions.} A natural research direction is to ask how broadly the memory-policy view of simulator-grounded design generalizes beyond fibers, whether the same closed-loop interplay between skill selection and skill evolution also holds for metasurfaces, integrated waveguides, and antenna structures, and whether the skills that emerge across these domains share recognizable physical motifs. A second open question concerns the dynamics of skill evolution itself, including how the designer's hard-case feedback shapes the long-term geometry of the skill space and under what conditions this evolution converges to a stable, transferable repertoire. We also view this work as an entry point toward bridging memory-augmented LLM agents and physics-aware scientific design, where simulator-grounded rewards offer a principled substrate for studying how reusable knowledge can be acquired, organized, and reused across long-horizon design processes.

\end{document}